%% file: final_version.tex
\pdfoutput=1
\PassOptionsToPackage{table}{xcolor}

\documentclass[11pt]{article}

\usepackage{acl}

\usepackage{times}
\usepackage{latexsym}

\usepackage[T1]{fontenc}
\usepackage[utf8]{inputenc}

\usepackage{microtype}

\usepackage{inconsolata}

\input{math_commands.tex}

\usepackage{hyperref} % hyperlinks
\usepackage{url} % simple URL typesetting
\usepackage{booktabs} % professional-quality tables
\usepackage{amsfonts} % blackboard math symbols
\usepackage{nicefrac} % compact symbols for 1/2, etc.
\usepackage{microtype} % microtypography
\usepackage{cleveref}
\usepackage{subcaption}
\usepackage{graphicx}
\usepackage{wrapfig}
\usepackage{amsmath,amsfonts,amsthm,amssymb}
\usepackage{mathtools}
\usepackage{multirow}
\usepackage{makecell}
\usepackage{comment}
\usepackage{enumitem}
\title{SAEs Are Good for Steering -- If You Select the Right Features}

\author{
Dana Arad$^{1}$\thanks{
Work partially done during an internship at Amazon.}  \quad Aaron Mueller$^{2}$  \quad Yonatan Belinkov$^1$  \\
$^1$Technion -- Israel Institute of Technology \quad $^2$Boston  University \\[1ex]
\href{mailto:danaarad@campus.technion.ac.il}{\texttt{danaarad@campus.technion.ac.il}} \quad \href{mailto:amueller@bu.edu}{\texttt{amueller@bu.edu}} \\
 \href{mailto:belinkov@technion.ac.il}{\texttt{belinkov@technion.ac.il}} 
}

\begin{document}
\maketitle

\begin{abstract}

Sparse Autoencoders (SAEs) have been proposed as an unsupervised approach to learn a decomposition of a model's latent space. 
This enables useful applications such as steering---influencing the output of a model towards a desired concept--without requiring labeled data. 
Current methods identify SAE features to steer by analyzing the input tokens that activate them.
However, recent work has highlighted that activations alone do not fully describe the effect of a feature on the model's output.
In this work, we draw a distinction between two types of features: \emph{input features}, which mainly capture patterns in the model's input, and \emph{output features}, which have a human-understandable effect on the model's output. 
We propose input and output scores to characterize and locate these types of features, and show that high values for both scores rarely co-occur in the same features.
These findings have practical implications: after filtering out features with low output scores, we obtain \mbox{2--3x} improvements when steering with SAEs, making them competitive with supervised methods.\footnote{Our code is available at \href{https://github.com/technion-cs-nlp/saes-are-good-for-steering}{https://github.com/technion-cs-nlp/saes-are-good-for-steering}.}

\end{abstract}

\begin{figure}[t]
 \centering
 \includegraphics[width=\linewidth]{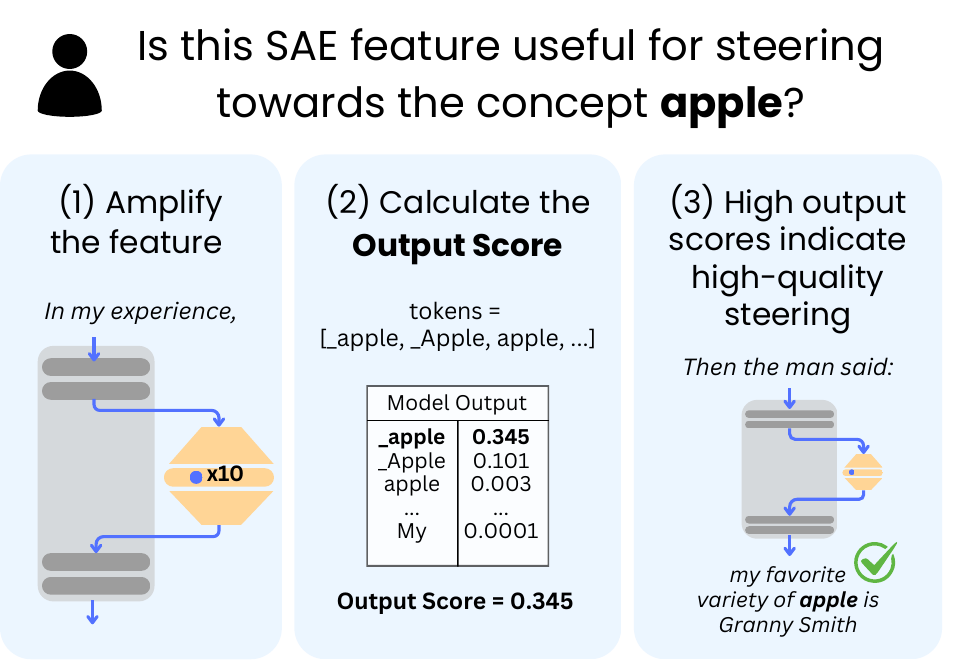}
 \caption{\textbf{Selecting features for steering.} (1) Given a concept to steer (''apple''), we amplify a candidate SAE feature during a single forward pass of the model on a neutral prompt. (2) We compute the feature's \textbf{output score} based on the rank and probability of representative after intervention. (3) Features with high output scores are  more likely to be effective for steering. 
}
 \label{fig:main}
\end{figure}

\begin{figure*}[t]
     \centering
    \begin{subfigure}{0.49\linewidth}
        \centering
        \includegraphics[width=\linewidth]{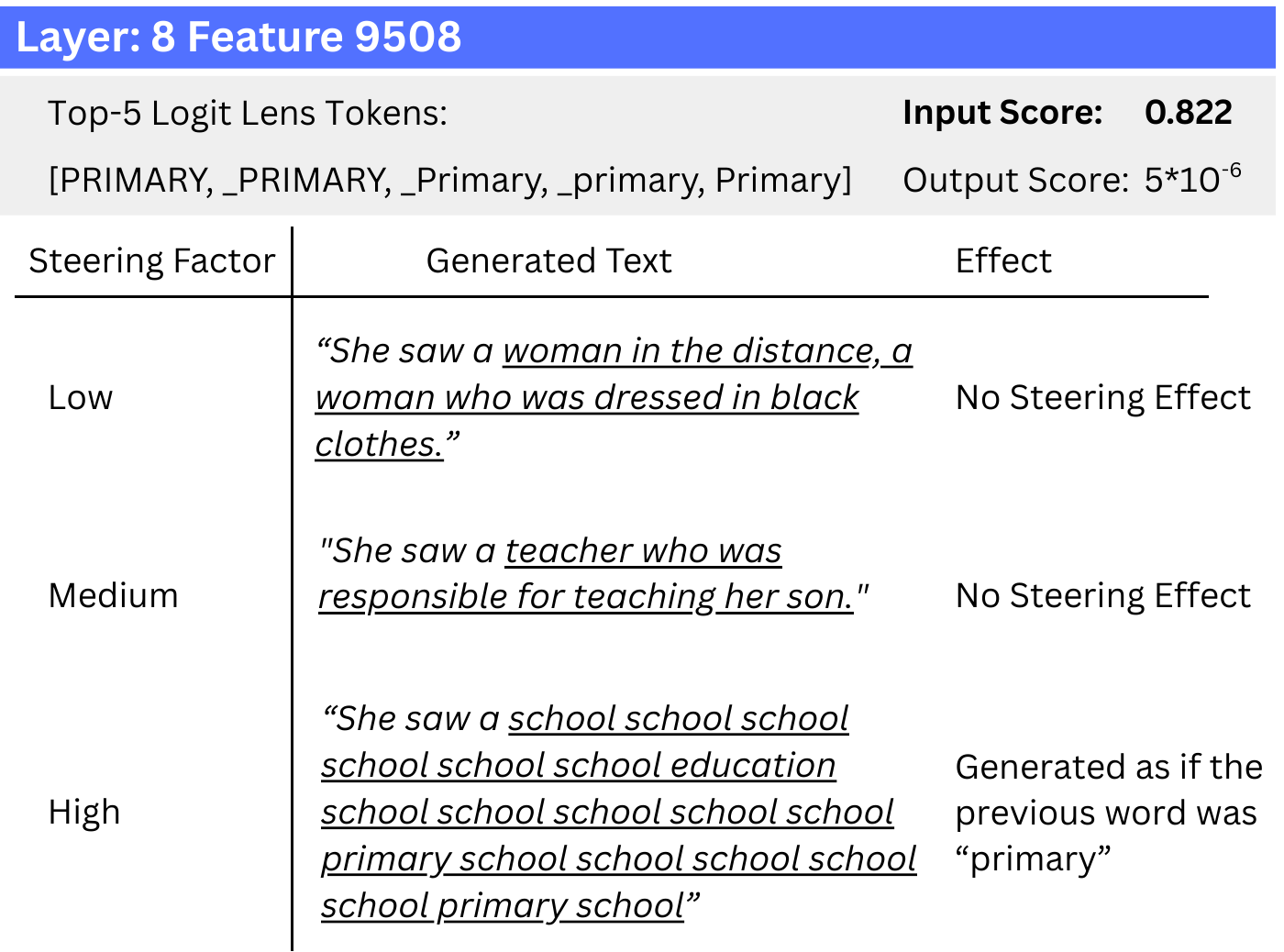}
        \caption{Steering with an input feature.} 
    \end{subfigure}
    \begin{subfigure}{0.49\linewidth}
        \centering
        \includegraphics[width=\linewidth]{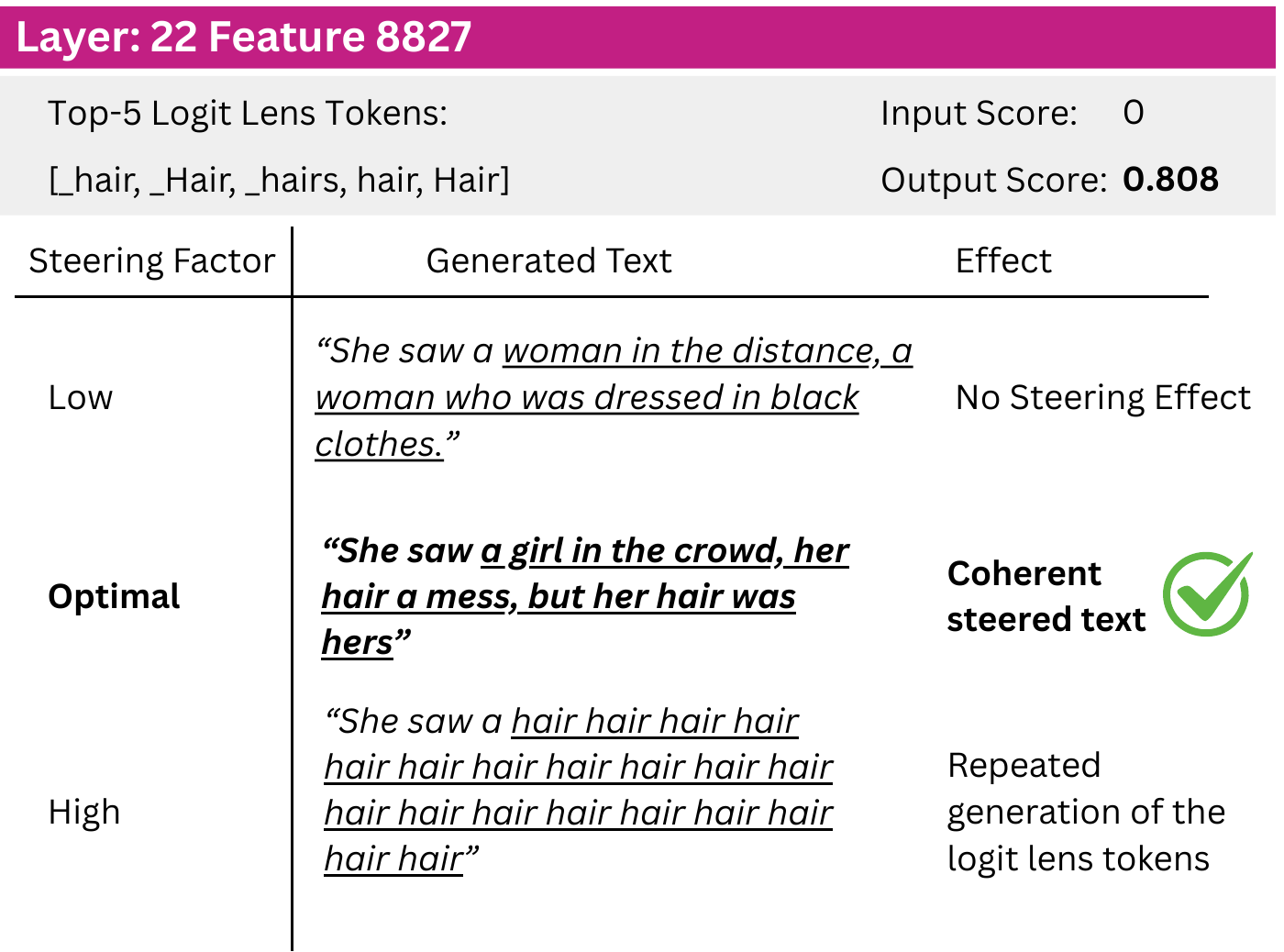}
        \caption{Steering with an output feature.} 
    \end{subfigure}
    
 \caption{\textbf{Examples of steering with input and output features.} \textbf{(a)} An input feature, which activates strongly on tokens like ''\_primary'' (leading to a high input score of $0.82$), fails to steer generation meaningfully; with a high steering factor, the model degenerates into repeating the token ''school'', as if continuing from the word ''primary''. \textbf{(b)} An output feature, with an output score of $0.81$, yields meaningful, coherent generations when steered at an optimal steering factor.}
 \label{fig:steering}
\end{figure*}

\section{Introduction}
Sparse autoencoders (SAEs) have shown promise in extracting human-interpretable features from the hidden states of language models (LMs) \cite{bricken2023monosemanticity, scaling2024templeton}. One appealing usage of SAEs is to enable fine-grained interventions such as generation steering \cite{o2024steering, durmus2024steering, marks2025sparse}. 
However, selecting the right features for intervention is an open problem.
Current approaches typically select features to steer based on their activation patterns, i.e., the input texts that most strongly activate a given feature \cite{interpretable2024huben}.
While input-based activations can reveal meaningful patterns, recent work highlights a critical limitation: a feature’s activations are not necessarily the same as its causal effect on the model’s output \cite{durmus2024steering, paulo2024automatically, gurarieh2025enhancing}.
As a result, the way features are selected can lead to suboptimal steering, reducing its consistency and reliability \cite{durmus2024steering}.

In this work, we formalize two distinct roles that features can play: \textbf{input features}, which capture patterns within the model’s input, and \textbf{output features}, whose main role is to directly influence the tokens the model generates.
To find them, we propose \textbf{input scores} and \textbf{output scores}. 
First, we obtain a representative set of tokens for each feature by applying the logit lens to SAE weights; this projects the weights directly into the vocabulary space \cite{nostalgebraist, saelogitlens}. 
We define input features as having high input scores---i.e., high overlap between their top-activating tokens and top logit lens tokens.
We define output features as having high output scores---i.e., intervening on the feature increases the probability of its top logit lens tokens in the final output distribution.
Notably, the input score can be computed in parallel for all features over a general dataset; the output score requires only one forward pass and no concept-specific data.
We quantitatively show that these roles rarely co-occur and tend to emerge at different layers in the model. 
Specifically, features in earlier layers primarily act as detectors of input patterns, while features in later layers are more likely to drive the model’s outputs, consistent with prior analyses of LLM neuron functionality \cite{lad2024mstages,marks2025sparse}.

By calculating the input and output scores of features extracted from Gemma-2 (2B and 9B) \cite{team2024gemma}, we show that features with high output scores are more effective for coherent and high-quality steering.
This yields a practical feature selection method, illustrated in \Cref{fig:main}: starting with a (typically small) set of candidate features for steering, our scores are computed over this set to select a more effective subset.
\Cref{fig:steering} demonstrates the difference when steering with input features vs.\ output features:
steering with a feature that has a high input score but low output score fails to meaningfully influence the generation.
In contrast, steering using a feature with high output score and low input score
yields better steering, as well as more fluent and semantically coherent completions.

We demonstrate the effectiveness of these insights on the recent AxBench \cite{wu2025axbench}, a benchmark for evaluating steering methods. While AxBench found SAEs to be poor for steering, our feature selection results in a 2--3x improvement, causing SAE steering (an unsupervised method) to score significantly closer to supervised methods like LoRA \cite{xu2024lora}.

In summary, our contributions are threefold:
\begin{itemize}[noitemsep,topsep=0pt]
  \item We propose a taxonomy of features according to whether they are more sensitive to analyzing the input or affecting the output, and propose ways of categorizing features into these different roles.
  \item We propose a practical method for finding features effective for steering. 
  \item Using our results, we engage with current debates on the utility of SAEs for steering, and characterize why these approaches did not find strong results.
\end{itemize}

\section{Preliminaries}

\subsection{Sparse Autoencoders}
Sparse Autoencoders (SAEs) were recently proposed as a method to address the problem of \textbf{polysemanticity}, where individual neurons entangle multiple unrelated concepts, and where a single concept may be distributed across many neurons \cite{bricken2023monosemanticity}.
Given a hidden representation $x \in \mathbb{R}^n$, an SAE consists of an encoder and a decoder, defined as:
\begin{align}
  a(x) &:= \sigma(W_{\text{enc}} x + b_{\text{enc}}), \\
 \hat{x}(a) &:= W_{\text{dec}} a + b_{\text{dec}}.
\end{align}
where $W_\text{enc}, W_\text{dec}, b_\text{enc}, d_\text{dec}$ are trainable parameters of the SAE.

The encoder maps the latent $x$ into a higher-dimensional sparse vector $a(x)$, or $a$ for short, which we refer to as the activations. The decoder reconstructs $x$ from $a$ as a sparse linear combination of the learned features, given by the columns of $W_{\text{dec}}$. 
Sparsity and non-negativity of the activations are enforced through the non-linearity $\sigma$, often JumpReLU \cite{rajamanoharan2024jumprelu}, and regularization.

\subsection{The Logit Lens}
The logit lens is a widely used interpretability tool for analyzing the hidden representations of language models \cite{nostalgebraist}. 
Given a hidden state $x \in \mathbb{R}^n$ at any layer of the model, the logit lens passes $x$ through the final layer norm, $\text{LN}$, then projects $x$ onto the vocabulary space by applying the unembedding matrix $W_{\text{unembed}} \in \mathbb{R}^{n \times |\mathcal{V}|}$, where $\mathcal{V}$ is the model's vocabulary. 
This produces a vector of predicted logits:
\begin{align}
 \ell(x) := W_{\text{unembed}}^\top(\text{LN}(x)).
\end{align}
The resulting logits $\ell(x)$ can be interpreted as the model's token predictions. We denote the top-$k$ predicted tokens as $\ell(x)_k$.

Recent work has demonstrated that the logit lens can be applied not only to hidden representations, but also model weights \cite{dar2023analyzing}, gradients \cite{katz2024backward}, and even in multi-modal settings \cite{toker2024lens}.

\citet{saelogitlens} suggested applying the logit lens to SAE feature weights as a way to interpret their roles. 
To interpret an SAE feature $f_i$, corresponding to $W^i_\text{dec}$, the $i$-th column in the decoder matrix, we compute:
\begin{align}
\ell(f_i) = W_{\text{unembed}}^\top \left( \text{LN}\left(W_{\text{dec}}^i\right) \right)
\end{align}

Following this body of work that views the logit lens as informative
explanations to models' computations and weights, 
we view $\ell(f_i)_k$ as a faithful explanation of the feature's role. We use $k=20$, and denote this list as $\ell$ for brevity.

\subsection{Steering LMs}
We define \textbf{steering} as influencing the output of an LM towards a desired concept. Successful steering should maintain the quality and coherency of the generated text. In other words, we seek a \emph{minimal} change to a model's computation that adds or subtracts a concept's influence.
Steering can be done by various methods, recently using SAEs \cite{durmus2024steering, wu2025axbench}. 

Formally, given a model $M$ and a prompt $x$, we obtain a steered text $\tilde{y}$ by applying an intervention $\Phi(\cdot)$ on some intermediate representation $h$:

\begin{align}
 \tilde{y} = M_{h \leftarrow \Phi(h)}(x)
 \label{eq:steering}
\end{align}

Similarly to \citet{scaling2024templeton}, in order to steer an LM towards a concept $c$ encoded in an SAE feature $f_{i}$
at layer $l$, we define $\Phi$ as follows:
first, we pass a prompt prefix $p$ through the model. At layer $l$, we pass the latent representation $x^l$ through the SAE encoder to obtain the activations vector, $a$. 
We record the max-activating feature, denoted $a_{\text{max}}$. 
Then, we obtain a new activation vector using steering factor $s$:
\begin{align}
 \tilde{a} = \begin{cases}
a_j & j \neq i\\
a_j + s \cdot a_{\text{max}} & j = i
\end{cases}
\label{eq:atilde}
\end{align}
We pass the steered activation vector through the SAE decoder, to obtain $\Phi(x^l) = W_{\text{dec}} \tilde{a} + b_{\text{dec}}.$
, and continue as usual with the rest of the forward pass. 

Similarly to \citet{makeloveevaluating}, to evaluate steering success we measure the generation success w.r.t.\ $\ell_k$ by calculating the number of appearances of any token in $\ell_k$ in the generated text. 
Given a set of sentences $S$:
\begin{equation}
\text{Gen Success@k}(S) = \frac{\sum_{s \in S}{|\{t \in s \mid t \in \ell_{k}\}|}}{|S|}
\end{equation}
Additionally, we use perplexity measured using Gemma-2-9B to quantify the generation coherence of the entire generated text.

\section{Feature Roles Across Layers}
\label{sec:roles}

In this section, we explore how features specialize across different layers by examining their relationship to the model’s input and output tokens. 
We first define input and output scores that measure the relationship of a feature with the model's inputs or outputs. 
Then, we describe experiments that report these scores across model layers. 

\subsection{Input Features}
An input feature is a feature whose behavior is closely tied to the tokens that activate it. Intuitively, if a feature consistently activates on a particular set of tokens, and its logit lens representation reflects the same tokens, then it is likely capturing information directly from the input.

\paragraph{Input Score.}
Given a large corpus, 
for each feature, let $S$ denote a set of sentences where the feature activated strongly on some tokens in each sentence.
For each sentence, we find the maximally activated token. Let $T$ denote the set of top activated tokens across all sentences in $S$. The input score is the fraction of the top activated tokens that are found in $\ell$, the top tokens when projecting the feature with the logit lens:
\begin{align}
S_{in} = \frac{|\{t \in T \mid t \in \ell\}|}{|T|} 
\end{align}

In practice, we use the pre-computed activations from Neuronpedia \cite{neuronpedia} to obtain the sentences $S$. 
We verify that $S$ has at least 20 sentences and take a maximum of 100 sentences per feature.

\subsection{Output Features}
Output features were first mentioned by \citet{paulo2024automatically} as features whose effect on the model’s output can be easily explained in natural language.
Natural language explanations predispose us to errors in both precision \emph{and} recall \citep{huang-etal-2023-rigorously}; therefore, given a target concept, we quantify steering quality as consistency between the set of logit lens tokens and the set of tokens that the steering operation promotes.

\paragraph{Output Score.}
To measure the effect of a feature on the model's output distribution, we perform an intervention during a forward pass and evaluate the change in the rank and probability assigned to tokens in $\ell$.
We first use a neutral prompt ($x=$ ''In my experience,'') to obtain the model's prior distribution over token ranks and probabilities. Then, we intervene on the feature's activation value using a large steering factor (we use $10$), as in \Cref{eq:atilde}. 

We record the ranks of the tokens in $\ell$ and their probabilities; we denote the token with the highest rank as $\ell^*$, its rank as $r(\ell^*)$, and its probability as $p(\ell^*)$. The output score is then the difference in rank-weighted probabilities between the original and counterfactual output distributions:
\begin{align}
 P(\mathcal{M}) = (1 - \frac{r(\ell^*,\mathcal{M})}{|V|}) p(\ell^*,\mathcal{M})\\
 S_{out} = P(\mathcal{M}_{h\leftarrow\Phi(h)}) - P(\mathcal{M})
 \label{eq:sout}
\end{align}
where $V$ is the model's vocabulary. If $x$ is a neutral prompt, then $S_{out} \propto P(\mathcal{M}_{h\leftarrow\Phi(h)})$, so we can compute $S_{out}$ quickly using a single forward pass by only computing the rank-weighted probability after the intervention. 
This score is robust to the specific choice of neutral prompt; see details in \Cref{app:neutral_prompt}. 

\subsection{Experimental Setup} 
We focus our analysis on Gemma-2 (2B and 9B) using the Gemma-Scope 16K SAEs \cite{team2024gemma, lieberum2024gemma}, Llama-3.1 8B with Llama-Scope SAEs \cite{grattafiori2024llama, he2024llama}, and Pythia-70m \cite{biderman2023pythia, interpretable2024huben}. 
Our analysis spans 100 features randomly sampled from each layer.
For Pythia-70m we limited our sampling to features with at least 10 recorded activations, since many features do not have any recorded activations on Neuronpedia.

\begin{figure}[t]
    \centering
    \begin{subfigure}{\linewidth}
        \centering
        \includegraphics[width=0.8\linewidth]{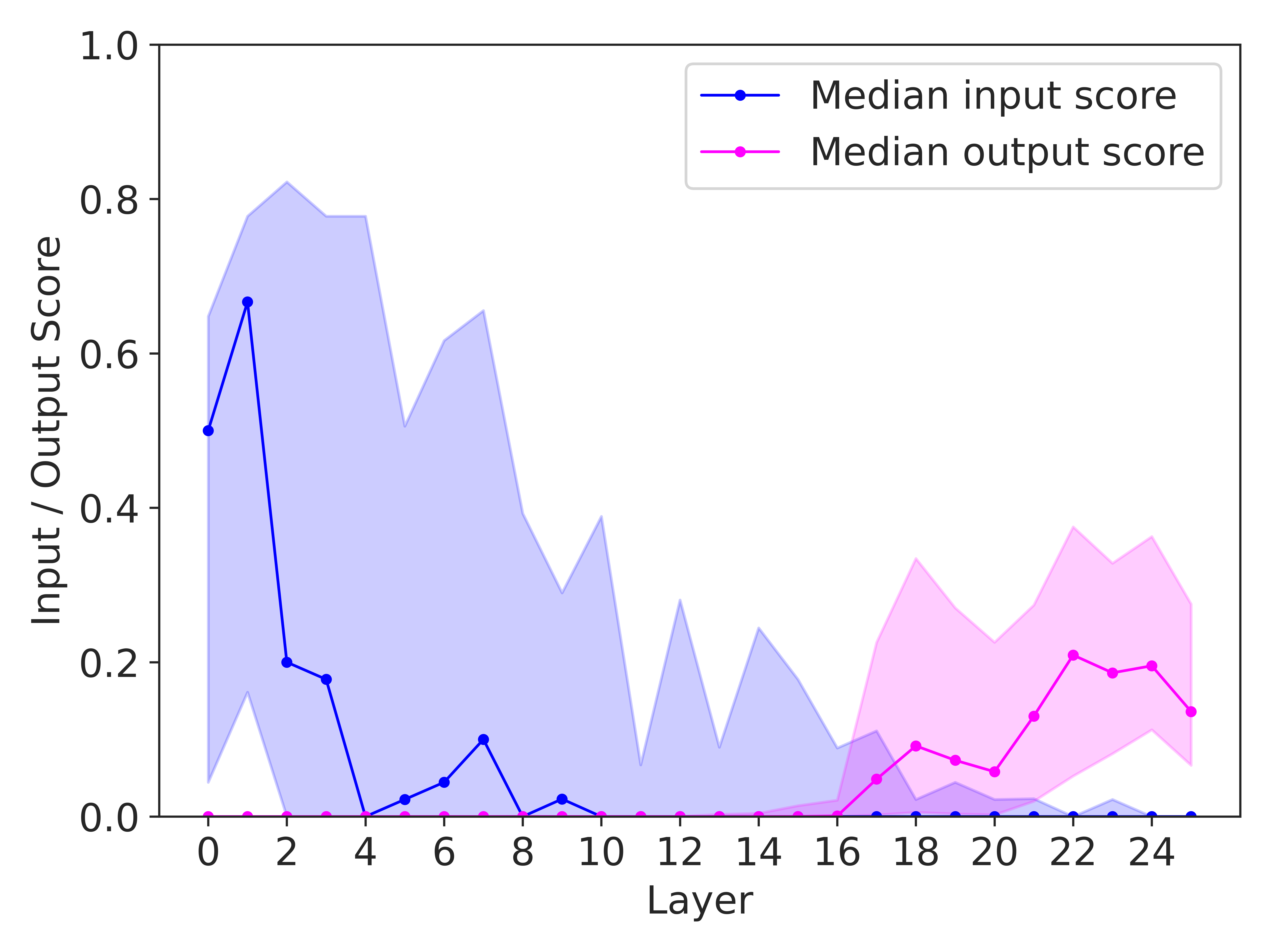}
        \caption{Gemma-2-2B.} 
    \end{subfigure}
    \vspace{0.5em} 
    \begin{subfigure}{\linewidth}
        \centering
        \includegraphics[width=0.8\linewidth]{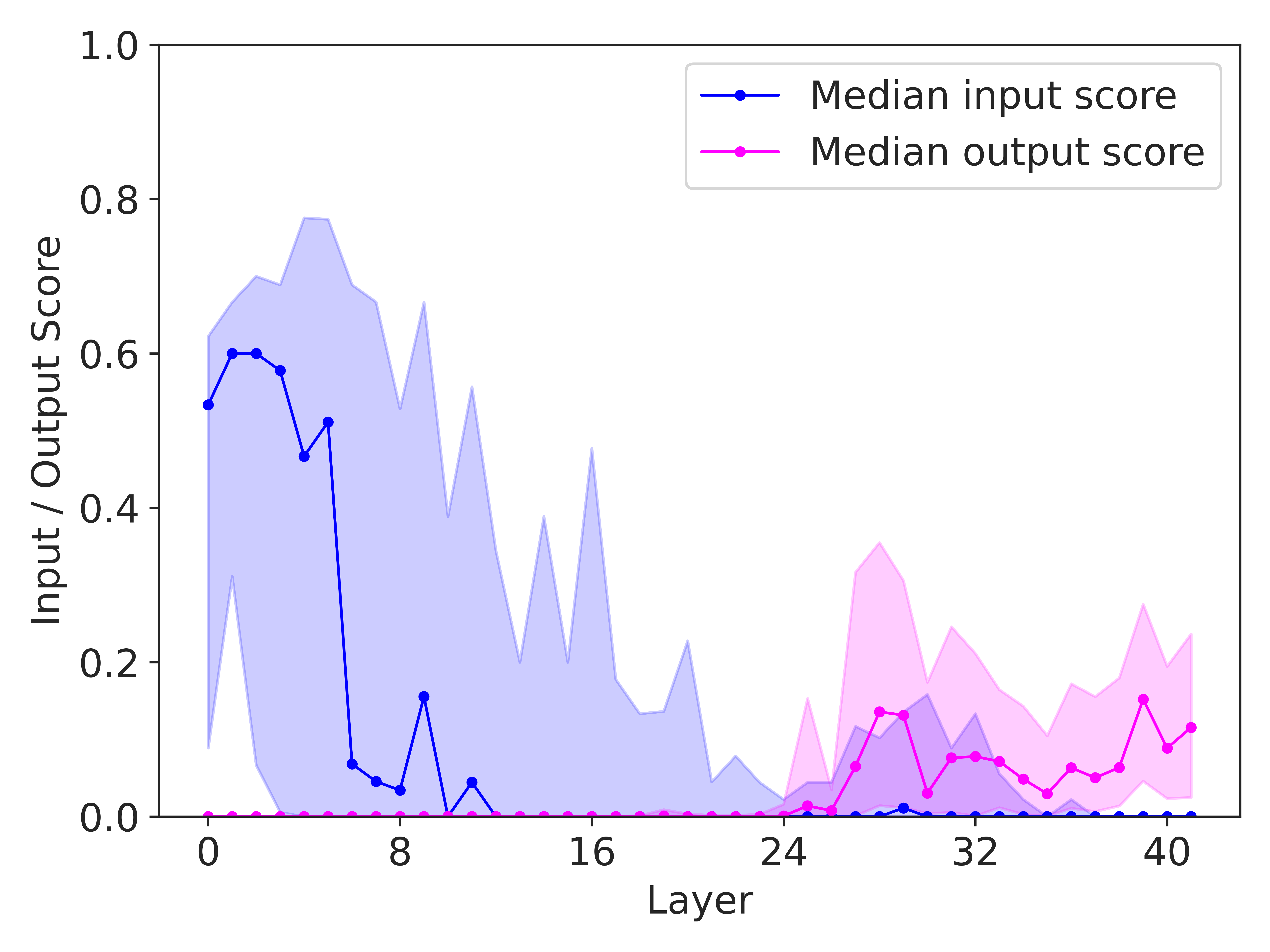}
        \caption{Gemma-2-9B.} 
    \end{subfigure}
    
    \caption{
 \textbf{Input and output scores across layers in Gemma-2-2B and Gemma-2-9B.}
 The solid lines represent the median \textcolor{blue}{input} score (blue) and \textcolor{magenta}{output} score (magenta), while the shaded regions denote the interquartile range (25th to 75th percentile), capturing the variability across features within each layer.
 Early layers are characterized by features with high input scores, while high output scores emerge in later layers.
  }
  \label{fig:roles}
\end{figure}

\subsection{Results}
\Cref{fig:roles} shows the distribution of input and output scores across layers for Gemma-2-2B and Gemma-2-9B. 
In early layers ($0$--$50$\% of model depth), features tend to have high input scores and near-zero output scores, suggesting they are predominantly input-aligned. Later layers ($66$--$100$\% of model depth) show an opposite trend: input scores drop to near-zero, while output scores increase significantly. These later-layer features no longer reflect the tokens they are activated on, but instead align with the tokens they promote in the model’s output, indicating a shift toward output-aligned behavior. 
Interestingly, middle layers exhibit low scores for both metrics, suggesting that these features may play intermediate roles that are neither purely input-aligned nor strongly output-promoting.

For Llama-3.1 we do not observe any trend in early layers of the model (\Cref{app:llama_pythia}), likely due to limitations of applying the logit lens to early layer representations \cite{nostalgebraist}. 
At around $50$\% of the model's depth we begin to observe non-zero values for both scores, with low values of input scores and increasingly growing values of output scores as layers progress, as in the Gemma-2 results.

For Pythia, we observe a slightly different trend (\Cref{app:llama_pythia}). 
While output score gradually increases around $50$\% of the model's depth as expected, the input score is mainly zero for most of the tested features. 
This may be due to the fact that this model is significantly smaller compared to the other models we examined ($70$ million parameters compared to $2$--$9$ billion), which may lead it to parse and encode information differently within its latent space.

Interestingly, and unlike Llama-3.1, early layers in both Gemma models and Pythia seem to be interpretable with the logit lens. 
We find that many features promote a coherent and human-understandable set of tokens, as reflected by high input scores as early as layers 0 and 1. (See \Cref{app:input_features} for examples.)

\section{Identifying Features for Steering}
\label{sec:steering}

The output score measures the alignment between the effects of SAE features on the model's output distribution and the expected set of tokens---in our case, their top logit lens tokens. 
In this section we hypothesize that features with high output scores are more effective for steering.
To test this, we evaluate generation success when filtering out features with low output scores at different thresholds. 

\begin{figure}[t]
 \centering
    \begin{subfigure}{0.49\linewidth}
        \centering
        \includegraphics[width=\linewidth]{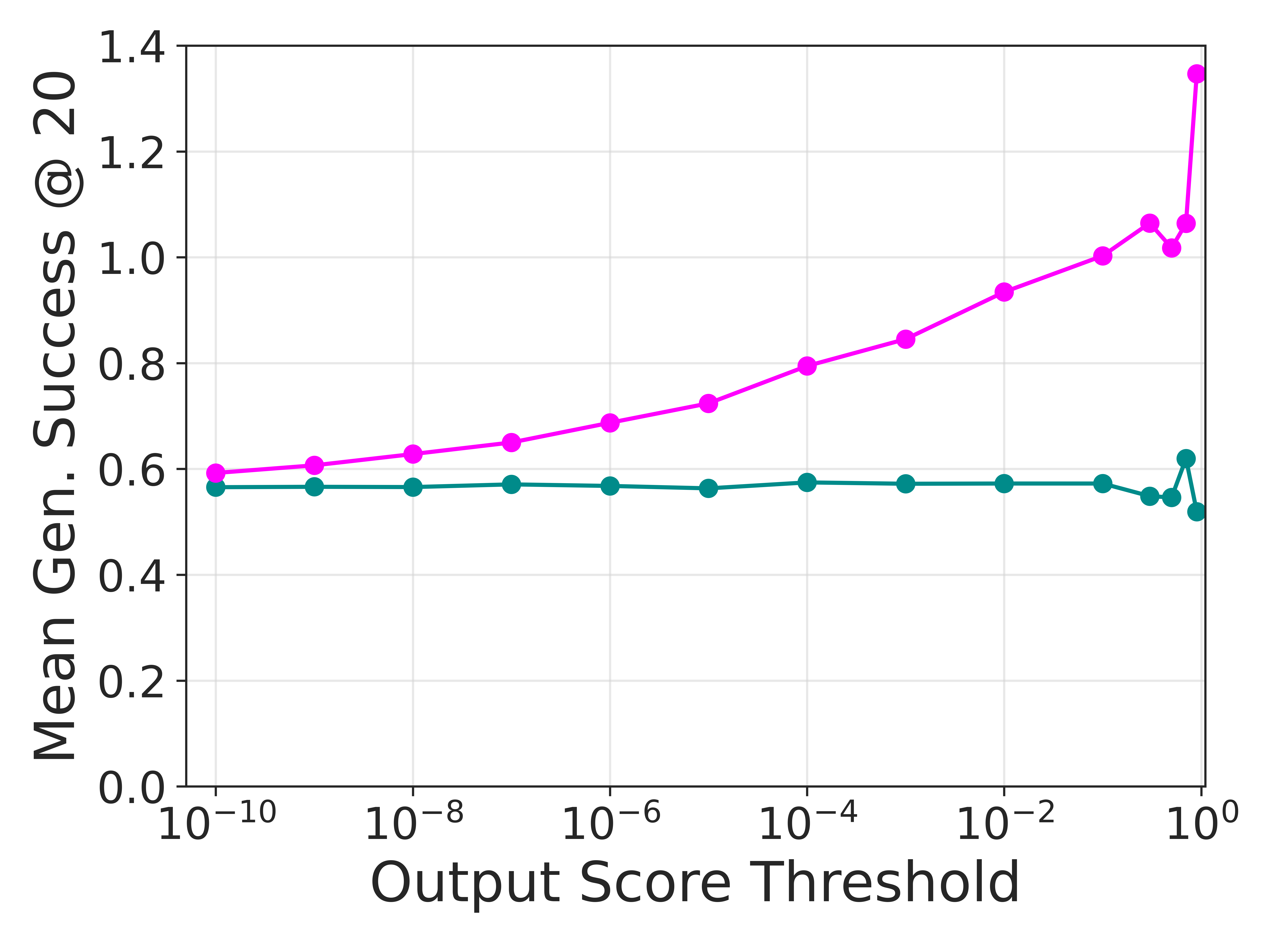}
        \caption{Gemma-2-2B.} 
    \end{subfigure}
    \begin{subfigure}{0.49\linewidth}
        \centering
        \includegraphics[width=\linewidth]{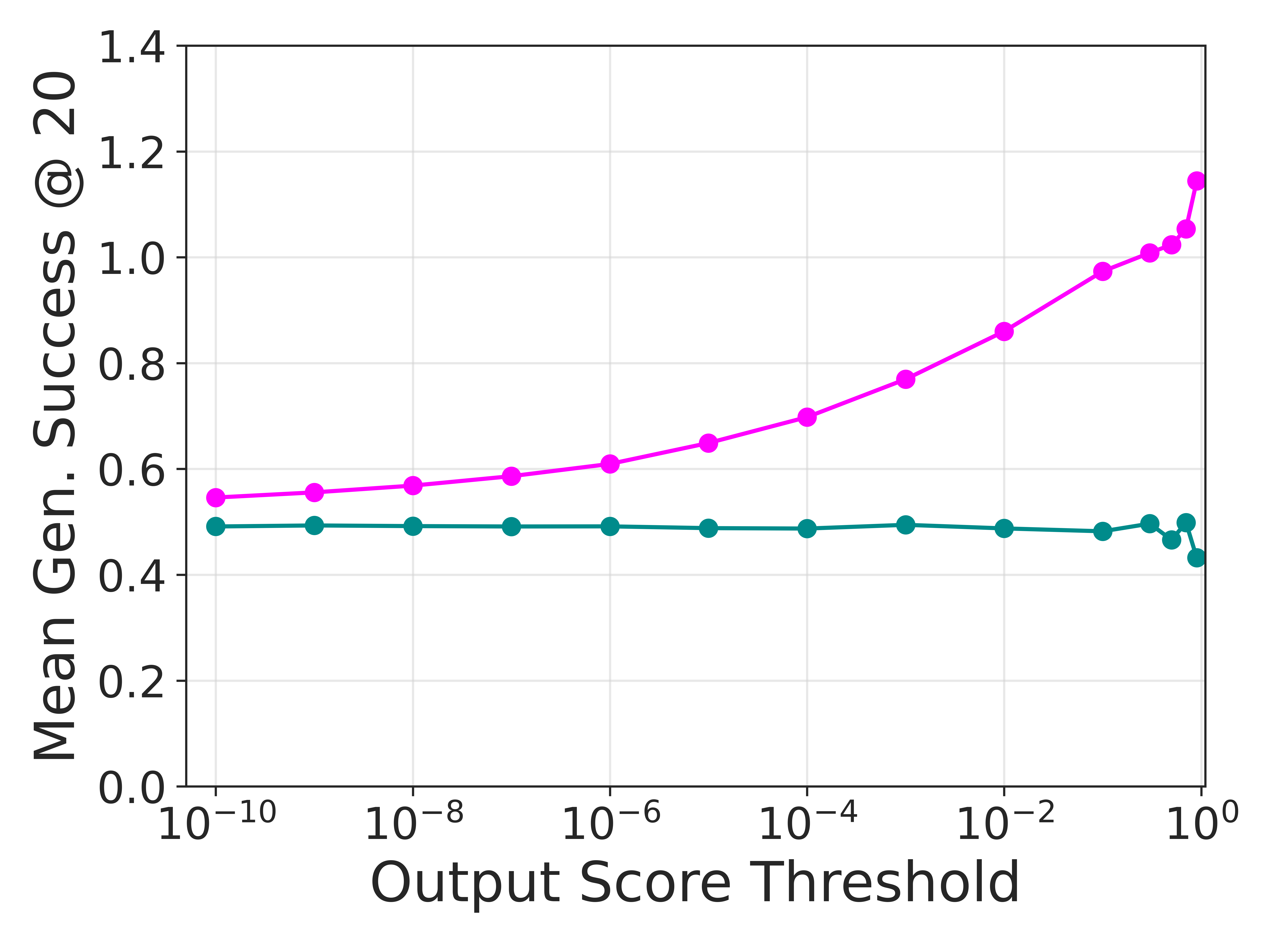}
        \caption{Gemma-2-9B.} 
    \end{subfigure}

 \caption{\textcolor{magenta}{Magenta} indicates the mean generation success@20 when filtering out features with output scores below different thresholds.
 \textcolor[HTML]{08A290}{Green} indicates the mean generation success@20 after filtering randomly sampled sets of features of the same size.
 Filtering results in significant increase in generation success.}
 \label{fig:os_th}
\end{figure}

\subsection{Experimental Setup}
\label{subsec:steering_setup}
We use 50 prompt prefixes and generate up to 20 tokens, obtaining 50 generated texts for each feature (more details in \Cref{app:steering}).
For each feature we calculate the mean generation success across the generated texts, and filter out steering factors leading to generation success greater than $3$. 
Intuitively, the generation success measures the rate in which the model generates concept-related tokens. 
Based on early experiments, we find that 3 is a good upper value that balances steering and coherence.
We choose the optimal steering factor as the one that maximizes $\frac{\text{Gen Success@20}}{\text{Perplexity}}$, where we normalize both metrics to a $0$--$1$ range by dividing with the maximum value across all data samples. 

\subsection{Qualitative Results}
\Cref{fig:steering} demonstrates steering with two features from Gemma-2-2B: one having a high input score and low output score (an input feature), and the other, an output feature, having a high output score and low input score.
Steering using the input feature fails to meaningfully influence the generation. When using a high steering factor, this results in repeatedly generating tokens related to the feature’s activation, as if continuing from the word ''primary''.
In contrast, steering using a feature with high output score and low input score using an optimal steering factor yields fluent and semantically coherent completions. Additional examples are shown in \Cref{app:steering}.

\begin{figure}[t]
 \centering
     \begin{subfigure}{0.49\linewidth}
        \centering
        \includegraphics[width=\linewidth]{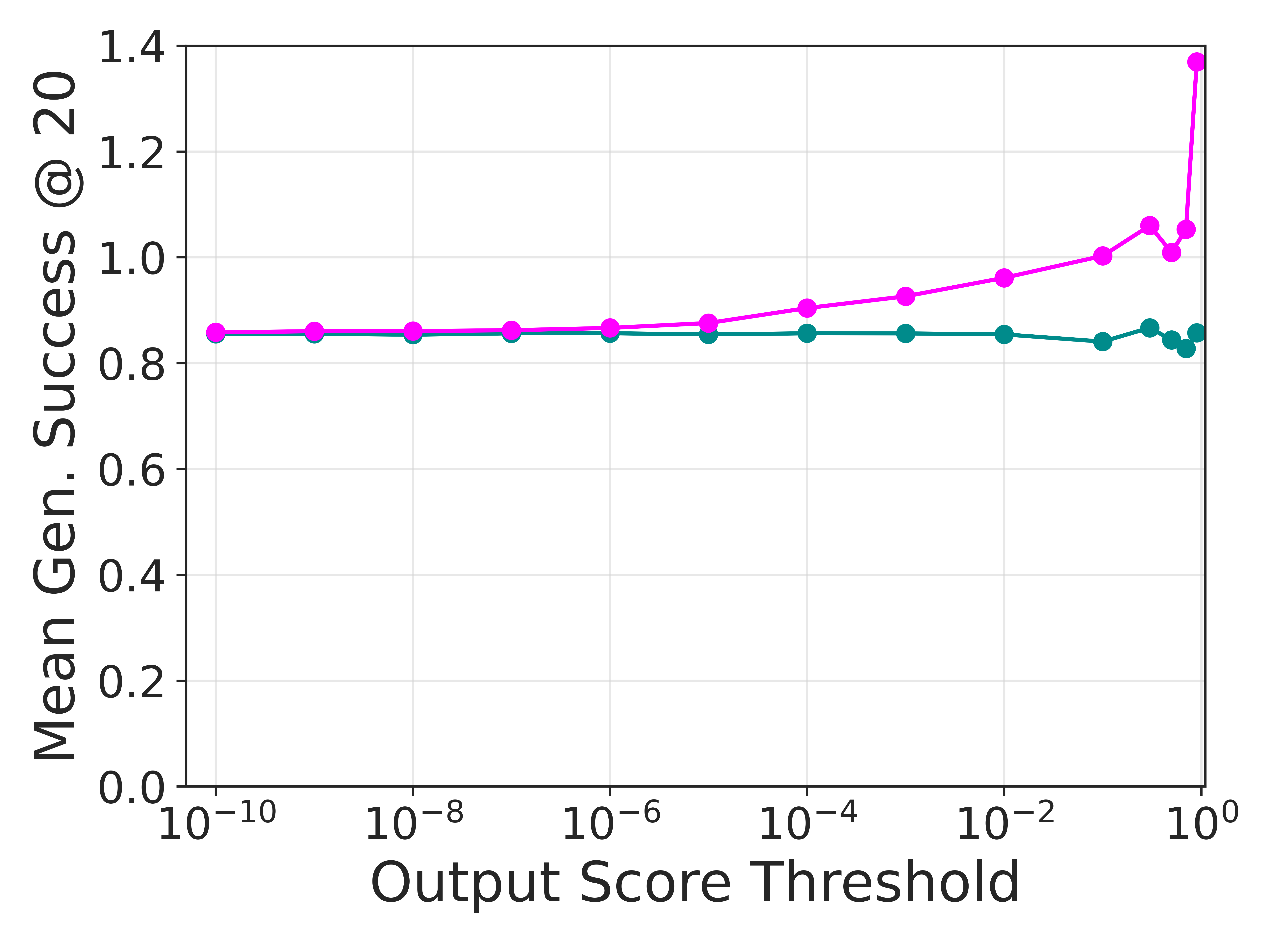}
        \caption{Gemma-2-2B.} 
    \end{subfigure}
    \begin{subfigure}{0.49\linewidth}
        \centering
        \includegraphics[width=\linewidth]{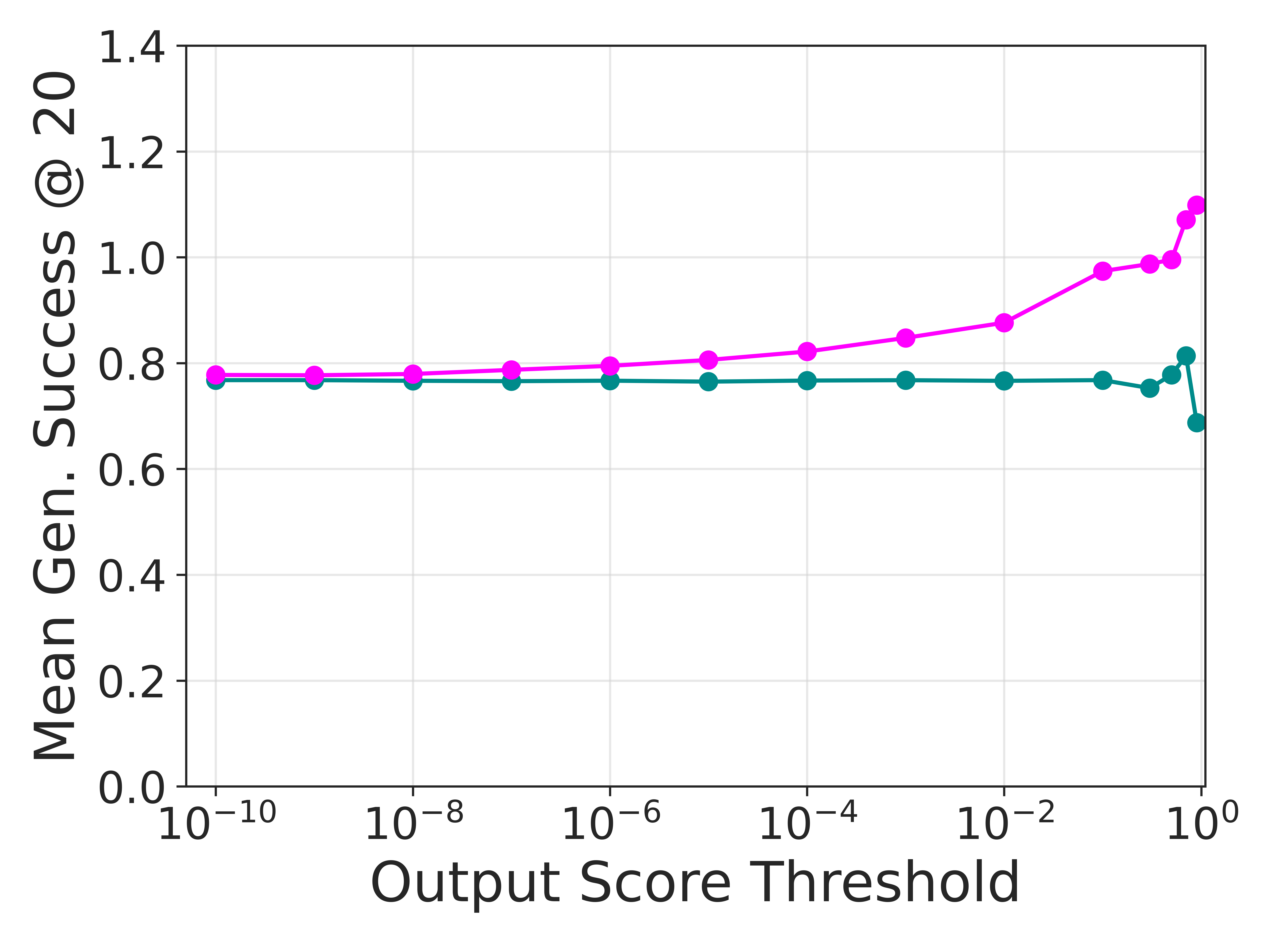}
        \caption{Gemma-2-9B.} 
    \end{subfigure}

 \caption{Even in later layers of the model (16--25 for Gemma-2-2B and 24--41 for Gemma-2-9B), filtering features with low output scores increases mean generation success.
\textcolor{magenta}{Magenta}: filtering by output scores. \textcolor[HTML]{08A290}{Green}: filtering random sets of features of the same size.}
 \label{fig:th_late_layers}
\end{figure}

\subsection{Quantitative Results}
\Cref{fig:os_th} shows the mean generation success @ 20 of steered generations when filtering out features with output scores below varying thresholds (\textcolor{magenta}{magenta}) on Gemma-2-2B and Gemma-2-9B. 
As the threshold increases, performance improves steadily, indicating that features with higher output scores consistently lead to more successful steering. 
We observe an increase in the mean generation success score from around $0.5-0.6$ for both models without any filtering, to $1.1$--$1.4$ using a threshold of $0.9$. A threshold of $0.01$ is sufficient for filtering out about $60\%$ of the features, increasing the mean generation success by around $0.4$ points. 
We compare this against a random baseline (\textcolor[HTML]{08A290}{green}): filtering randomly sampled subsets of features of the same size does not lead to any significant improvements (results are average of 10 random samples per subset size). 
Llama-3.1 and Pythia show similar trends; see \Cref{app:llama_pythia}.

The results in \Cref{sec:roles} suggest that features with high output scores occur predominantly in later layers. 
\Cref{fig:th_late_layers} shows the generation success when filtering based on the output score, evaluated only on features from later layers of the model: 16--25 for Gemma-2-2B and 24--41 for Gemma-2-9B.
Taking only features from later layers, \mbox{Gemma-2-2B} and Gemma-2-9B achieve generation success scores of about $0.8$.
By considering only top-scoring features, the mean generation success increases to around $1.1$--$1.4$ for both models. 
These results show that even within these later layers, the output score is a useful tool for filtering out features that lead to poor steering results. 

\begin{table}[]
\small
\begin{tabular}{llcc}
 \toprule
  &  & \multicolumn{2}{c}{Gemma-2-9B-it} \\
  \cmidrule(lr){3-4}
  &  & \multicolumn{1}{c}{L20} & \multicolumn{1}{c}{L31} \\
\midrule
Our results &
$\text{SAE}_{(S_{out} \geq 0.1)}$ & 0.546 & \underline{0.470} \\
  & $\text{SAE}_{(S_{out} \geq 0.01)}$  & 0.338 & 0.454 \\
  & $\text{SAE}_{(S_{out} \geq 0.001)}$ & 0.373 & 0.415 \\
  & $\text{SAE}_{(S_{out} \geq 0.0001)}$  & 0.325 & 0.401 \\
  & $\text{SAE}_{(\text{No Filter})}$  & 0.293 & 0.387 \\
\midrule
\multirow{3}{*}{\begin{tabular}[c]{@{}l@{}}AxBench \\
reported \\
results \\ (Wu et al. 2025)\end{tabular}} & \cellcolor[HTML]{f0f0f0} Prompt & \cellcolor[HTML]{f0f0f0} \textbf{1.075}  & \cellcolor[HTML]{f0f0f0} \textbf{1.072}  \\
  & \cellcolor[HTML]{f0f0f0} LoReFT & \cellcolor[HTML]{f0f0f0} 0.777 & \cellcolor[HTML]{f0f0f0} 0.764 \\
  & \cellcolor[HTML]{f0f0f0} LoRA & \cellcolor[HTML]{f0f0f0} 0.602 & \cellcolor[HTML]{f0f0f0} 0.580 \\
  & ReFT-r1 & \underline{0.630} &  0.401 \\
  &  DiffMean &  0.322 &  0.158 \\
  &  SAE  &  0.191 &  0.140 \\
  &  SAE-A &  0.186 &  0.143 \\
\bottomrule
\end{tabular}
\caption{\textbf{Results on the Concept500 dataset from AxBench on instruction-tuned Gemma-2-9B.} (Top) Results when steering with SAEs after filtering out features with $S_{out}$ lower than different thresholds. (Bottom) Results reported by \citet{wu2025axbench}. \textbf{Bold} indicates the best score, \underline{underline} indicates the best score among representation-based methods. \colorbox[HTML]{f0f0f0}{Grey} indicates non-representation-based methods.
% Note that in our experiments, SAEs without any filtering achieve a higher score than reported in AxBench; see Footnote~\ref{footnote:axbench}.
After filtering based on output scores, SAEs achieve the best score among representation-based methods at L31, and reach 90.7\% of the best method's performance at L20.}
\label{table:axbench}
\end{table}

\subsection{Evaluation on AxBench}
AxBench was recently proposed as a dataset to evaluate steering methods \cite{wu2025axbench}. 
They compare steering with SAE features to other methods (including supervised methods), and find SAE features relatively ineffective.
However, we believe this is partially due to a non-principled selection of SAE features; we propose to remedy this using the output score.

We evaluate our findings on instruction-tuned Gemma-2-9B using the Concept500 dataset of \citet{wu2025axbench}, which includes 1000 $\langle$concept, SAE feature$\rangle$ pairs from layers 20 and 31 of the model. 
As in AxBench, we randomly sample 10 instructions for each concept-feature pair from instruction datasets aligned with the concept’s genre.
Five are used to select the optimal steering factor (as detailed in \Cref{subsec:steering_setup}), and the remaining five are used exclusively for evaluation.
We evaluate the steered texts using the metrics defined by \citeauthor{wu2025axbench}: (1) the concept score measures if the concept was incorporated in the generated text; (2) the fluency score measures the coherency of the text; and (3) the instruct score measures the alignment of the generated text with the given instructions. 
For each feature, we compute the harmonic mean of the three metrics. 
See \Cref{app:axbench} for additional details on the steering setup, metrics, and baseline methods.

\begin{figure}[t]
    \centering
    \begin{subfigure}{0.49\linewidth}
        \centering
        \includegraphics[width=\linewidth]{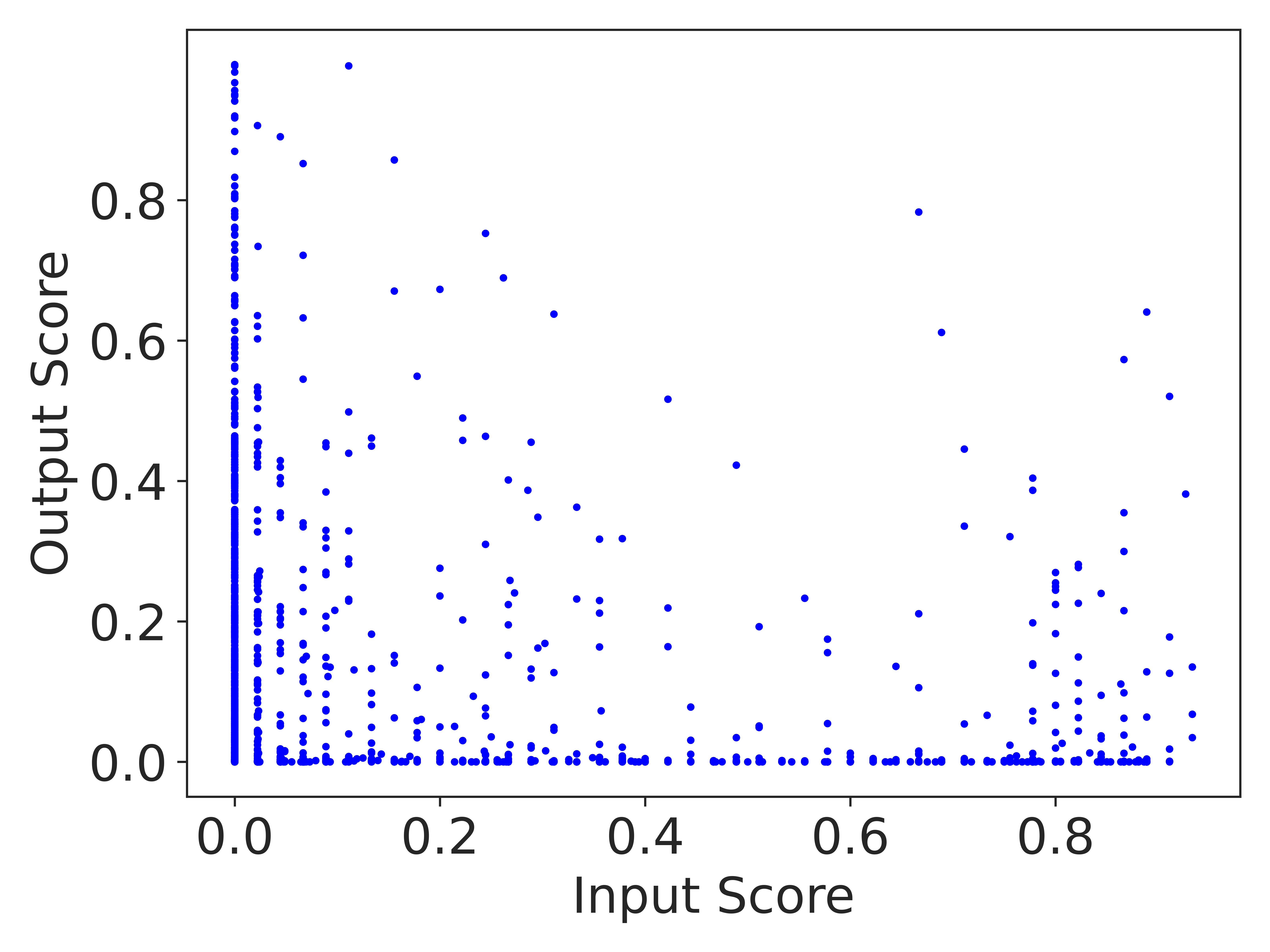}
        \caption{Gemma-2-2B.} 
    \end{subfigure}
    % \vspace{0.5em} 
    \begin{subfigure}{0.49\linewidth}
        \centering
        \includegraphics[width=\linewidth]{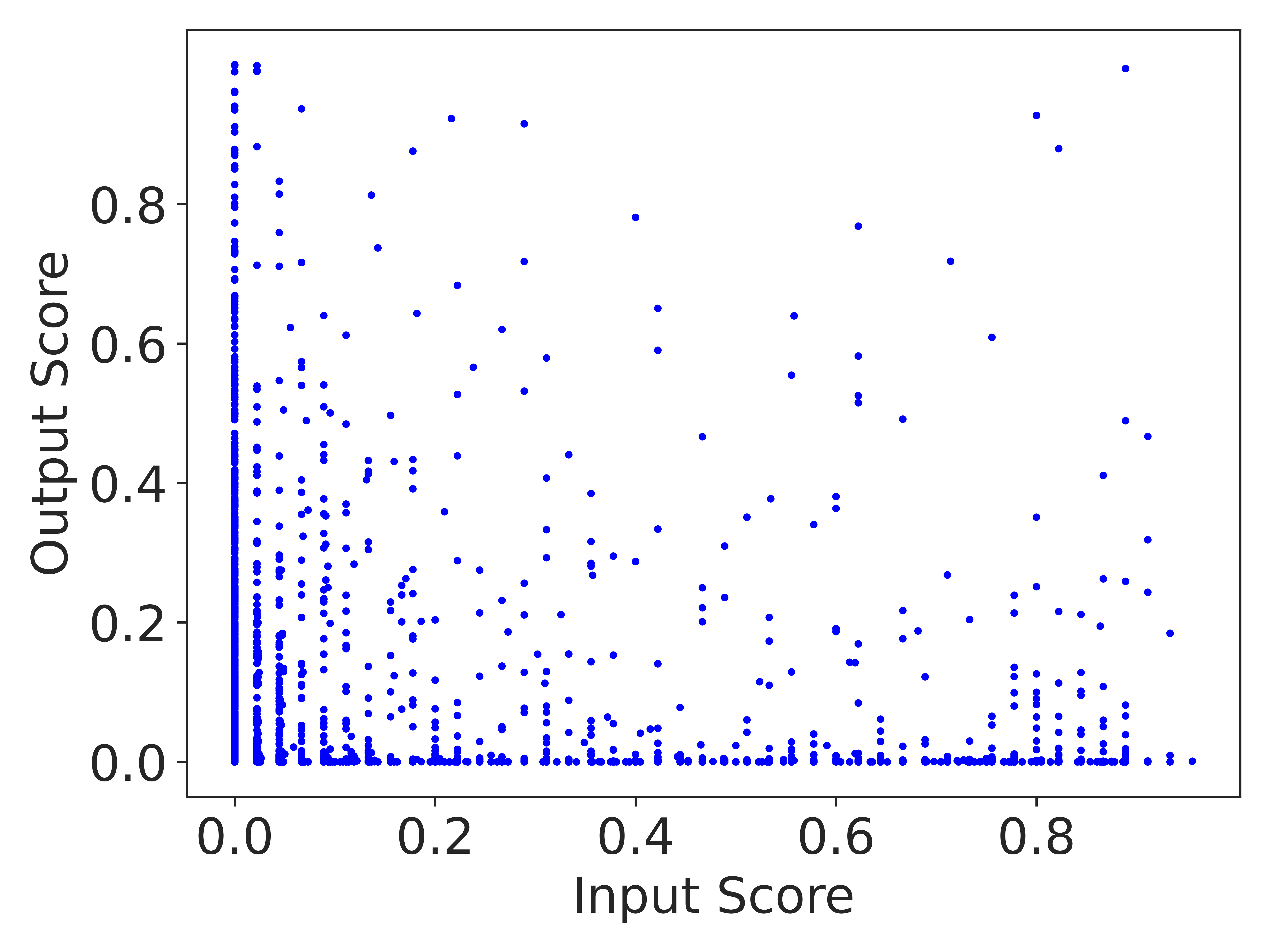}
        \caption{Gemma-2-9B.} 
    \end{subfigure}
    
    \caption{
\textbf{Relationship between input and output scores for features in Gemma-2-2B and Gemma-2-9B.} Most features lie near the axes, indicating that most features are either input \emph{or} output features---though a few are both.}
    \label{fig:io_scores}
\end{figure}

\begin{figure*}[ht]
 \centering
 \includegraphics[width=\linewidth]{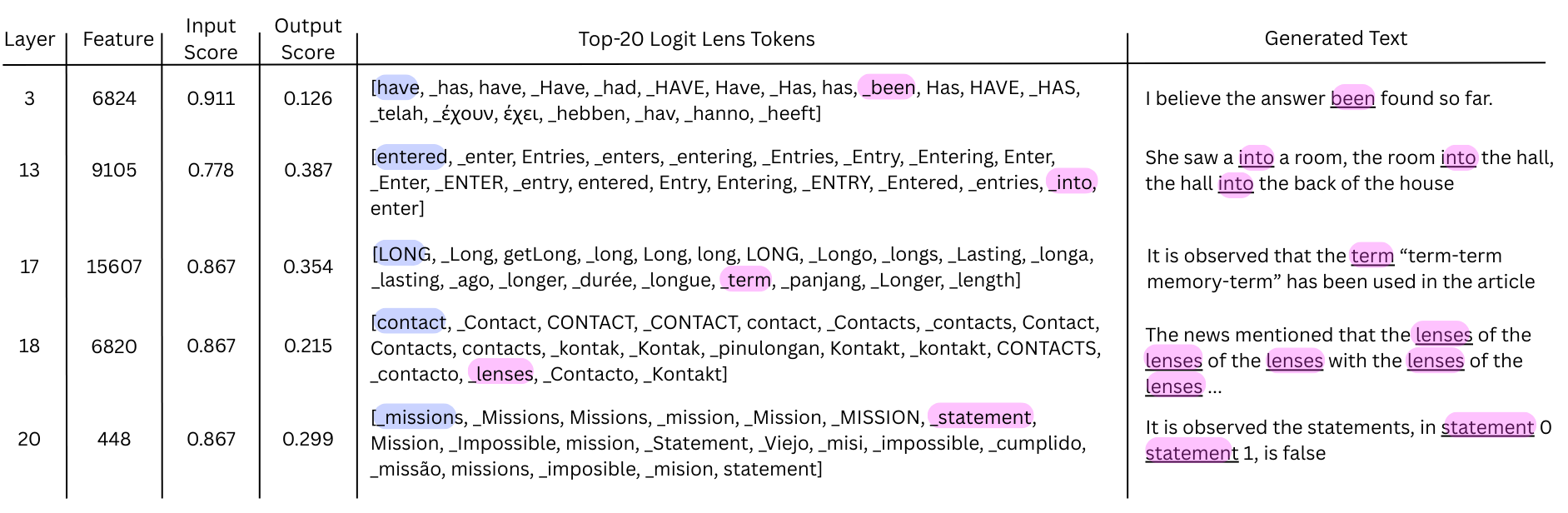}
 \caption{Generation results when steering with features that have both high input and high output scores in Gemma-2-2B. The top logit lens tokens (left; top-1 in \textcolor{blue}{blue}) do not appear directly in the generated text (right). The steered tokens (\textcolor{magenta}{magenta}), appearing lower in the logit lens ranking, often have strong collocational associations with the top logit lens tokens.}
 \label{fig:io_steering}
\end{figure*}

We report the mean score on layers 20 and 31 of instruction-tuned Gemma-2-9B (\Cref{table:axbench}). 
We find a nearly threefold improvement in SAE steering scores relative to \citeauthor{wu2025axbench}. 
Note that our replication of their experimental setting (without filtering) yields higher scores for SAEs; this can be a result of different sampled instructions per feature,\footnote{\citet{wu2025axbench} sample 10 instructions per feature from pre-existing datasets, but do not release these instructions. We sample 10 instructions from the same datasets, which may be different compared to the sample of \citeauthor{wu2025axbench}.\label{footnote:axbench}}
or possibly due to evaluation instability introduced by the use of an external LLM.
With output score filtering, SAE features top steering performance among the representation-based methods at L31; for L20, they get 90.7\% of the performance of the best method. 
This is in contrast with the results of \citeauthor{wu2025axbench}, where SAEs significantly underperforms ReFT-r1, a weakly supervised method they propose as a competitive alternative to prompting. 
These results demonstrates that with effective feature selection SAEs are comparable with existing methods, including supervised or weakly-supervised methods which require concept-specific datasets.

\section{The Relationship Between Input and Output Scores}

We next examine how input and output scores interact. 
\Cref{fig:roles} suggests that high input and output scores are rarely observed in the same layers, but do they sometimes co-occur in the same features?

\Cref{fig:io_scores} visualizes the relationship between the two scores: Indeed, most samples cluster near the axes, exhibiting either a high output score with a near-zero input score, or vice-versa.
However, there do exist hybrid features: features with both high input and high output scores.

\Cref{fig:io_steering} illustrates generation results when steering with  hybrid features from Gemma-2-2B.
Generated tokens that also appear in the feature's top-20 logit lens tokens are highlighted in \textcolor{magenta}{magenta}.
These are often not the top-ranked tokens under the logit lens, but they tend to rank moderately high and collocate with the top tokens (highlighted in \textcolor{blue}{blue}). For instance, for feature 6820 in layer 18, ``contact'' is the top logit lens token, but the output text repeatedly includes ``lenses'',
a token that appears further down the list but that is semantically and syntactically related. 

\begin{figure}
    \centering
    \begin{subfigure}{0.49\linewidth}
        \centering
        \includegraphics[width=\linewidth]{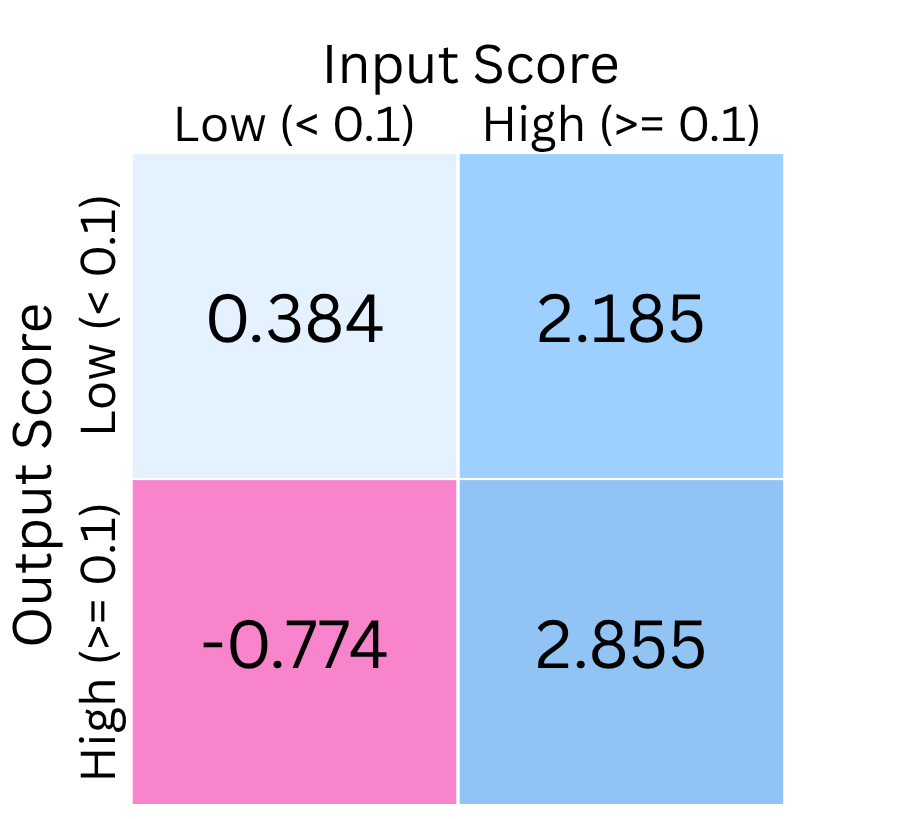}
        \caption{Gemma-2-2B.} 
    \end{subfigure}
    \begin{subfigure}{0.49\linewidth}
        \centering
        \includegraphics[width=\linewidth]{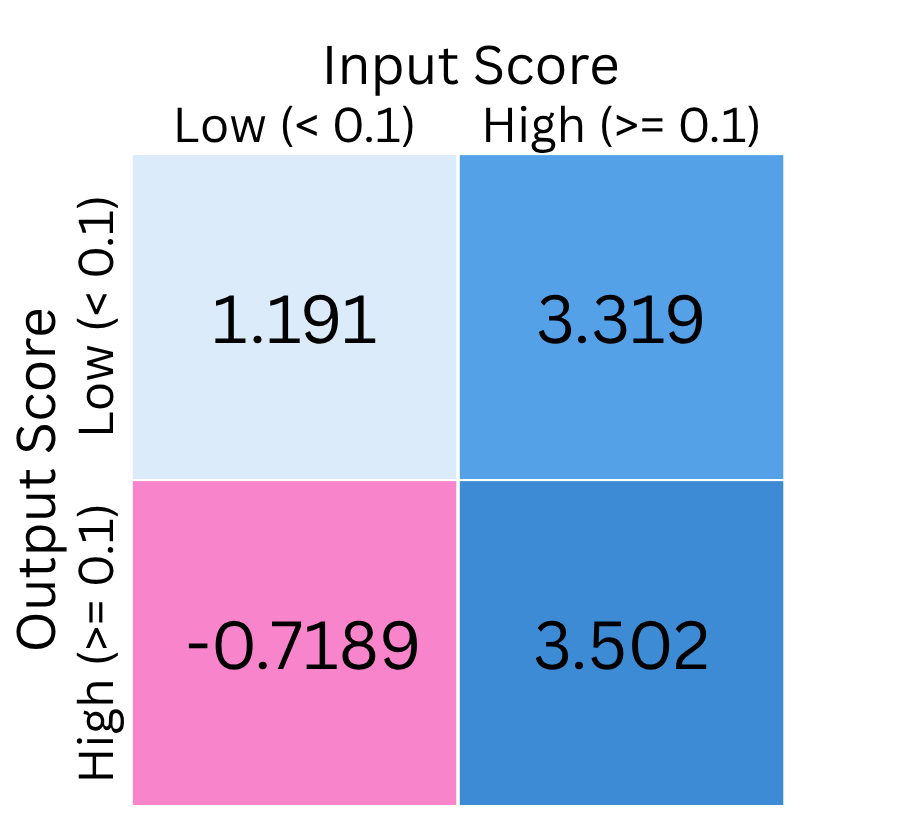}
        \caption{Gemma-2-9B.} 
    \end{subfigure}

 \caption{Pointwise Mutual Information (PMI) between generated tokens and top-1 logit lens tokens, grouped by input/output score thresholds. Features with high output scores and low input scores (\textcolor{magenta}{magenta}) tend to have negative PMI values, likely due to exact token repetition during generation, which indicates successful steering. In contrast, features with high input scores (\textcolor{blue}{blue}) consistently yield higher PMI values, indicating stronger collocational relationships with their top tokens.}
 \label{fig:pmi}
\end{figure}

We verify this intuition by quantifying collocation patterns between generated tokens and top logit lens tokens using their pointwise mutual information (PMI)\footnote{We use the pre-computed PMI over the webtext corpus in NLTK \cite{bird-loper-2004-nltk}, and only include token pairs that have this pre-computed score (100--200 pairs per model).}.
A PMI of zero indicates that two tokens co-occur no more frequently than chance, and negative and positive scores indicate lower- or higher-than-chance co-occurrence, respectively.

Features with high output score ($S_{out}	\geq 0.1$) and low input score ($S_{in} < 0.1$) have a negative PMI value on average (\Cref{fig:pmi}). 
This can be attributed to the high generation success of these features, i.e., that the top logit lens token is equal to the generated token. 
In most cases, two instances of the same token are not likely to consecutively co-occur, thus obtaining a low PMI score.

Importantly, features that have a high input score ($S_{in} \geq 0.1$) tend to generate tokens with significantly higher PMI relative to their top logit lens tokens, regardless of their output score values. 
This suggests that
hybrid features may be less favorable for steering, despite their high output score.

\section{Related Work}
\subsection{Stages of Processing in LMs}
The different stages of processing within NLP models have long been studied \cite{belinkov2017translation, zhang2018language, liu2019linguistic, brunner2020identifiability}. 
In transformer-based LMs, a large body of work shows that different properties emerge in different layers. Early layers focus on syntactic tasks such as POS, while semantic information appears in later layers \cite{bert2019tenney, elazar2021amnesic, geva2021keyvalue}. 
More recent work has demonstrated that intermediate layers are responsible for retrieving factual knowledge and enriching latent representations  \cite{meng2022locating, geva2023recall, hernandez2024linearity, arad2024refact}.

Another line of work has focused on so-called
prediction neurons, which increase the probability of coherent sets of tokens. This  work characterizes prediction neurons by properties of their logit lens distribution \cite{gurnee2024universal, lad2024mstages, saelogitlens}. In contrast, our output score directly measures the \emph{causal} effect of a feature on predicting a pre-defined set of tokens via counterfactual interventions. 
Relatedly, \citet{lad2024mstages} have shown that neurons in early layers pay more attention to input tokens in their proximity compared to later layers, while prediction neurons emerge later, after about 50\% of the model depth, in line with our findings on SAEs.

A closely related line of work aims to explain SAE features in natural language---for instance, by feeding inputs and activations into an external LLM \cite{bills2024explaining, interpretable2024huben}. 
However, this method results in errors in both precision and recall \citep{huang-etal-2023-rigorously}, and \emph{negatively} (albeit weakly) correlates with their causal role on average \citep{paulo2024automatically}. 
\citet{gurarieh2025enhancing} suggest that SAE features are better explained in terms of their activations \emph{and} their effect on the output(as quantified by projecting features into vocabulary space).
In this work, our aim is not to \emph{explain} features, but rather \emph{categorize} them with respect to their usefulness for steering. 
Additionally, we differentiate between two key feature roles (often mutually exclusive); this helps explain these prior findings and failure cases. 

\subsection{Steering LMs}
Many approaches exist for precisely influencing the outputs generated by LMs \cite{zou2023representation}.
These include prompt engineering \cite{wu2025axbench, taveekitworachai2024null}, steering vectors \cite{subramani2022extracting, teehan2022emergent, liu2023context} or
inference-time interventions on activations \cite{turner2023steering, van2024extending, rimsky2024steering}.
Steering was shown to be useful not only for directing generated content to a specific topic or concept, but also for style transfer \cite{lai2024style}, mitigating hallucinations \cite{li2023inference, simhi2024constructing}, and debiasing \citep{li2025fairsteer}. 

While most work in this area steers via interventions on full hidden states, earlier work attempted to influence model behavior by intervening on small sets of neurons \cite{bau2018identifying}. 
However, the polysemanticity of neurons makes them poor candidates for effective steering \cite{bricken2023monosemanticity}. 
In contrast, SAEs were shown to result in meaningful steering towards human-understandable concepts; a famous example involved steering toward responses related to the Golden Gate Bridge \cite{scaling2024templeton}, and another involved amplifying or mitigating social and political biases \cite{durmus2024steering, marks2025sparse}.
Recently, \citet{wu2025axbench} evaluated SAE steering against many methods, including supervised methods such as full fine-tuning, prompting, and difference-in-means \cite{pmlr-v48-larsen16}. 
They found that even these simple baselines outperform SAEs.
However, our work shows that most of the gap can be closed via more careful choice of SAE features. 
Typical work chooses features for steering based on natural language explanations generated based on each feature's activation patterns \cite{interpretable2024huben, durmus2024steering};
our findings instead suggest that \emph{influence on output} is a better proxy for steering efficacy, and that input activations have little predictive power for finding good steering features.

\section{Conclusions}
We have formalized and analyzed two roles demonstrated by sparse autoencoder (SAE) features.
We have defined the notion of an input score, which captures the alignment of a feature's activations with its top logit lens tokens, and an output score, which quantifies the alignment of the top logit lens tokens with the feature's effect on the model's generations.
We demonstrate that features with high output scores are significantly more effective for steering, whereas features with high input scores are relatively ineffective, even when they appear relevant to the steering concept.

\section*{Limitations}
While our work provides an efficient framework for identifying and leveraging output features for generation steering, several limitations remain.
First, our analysis is restricted to features extracted from the residual stream, and does not account for features derived from other components such as attention or MLP layers. As a result, our taxonomy may not capture the full range of functional roles present across the model.

% Following previous work, we focus our analysis on the Gemma-2 family of models and the Gemma-Scope SAEs \cite{team2024gemma, lieberum2024gemma}. While these models are widely adopted for interpretability research, further analysis across different LMs and SAE architectures would be beneficial to assess the robustness of our method.

Additionally, our method focuses on steering using a single SAE feature.
In practice, interactions between features may lead to better and more complex effects on generation \citep{wattenberg2024relational,singhvi-etal-2025-using}. 
Understanding how multiple features combine or interfere remains an open challenge.

\section*{Ethical Considerations}
Our work suggests a framework that improves one's ability to choose meaningful SAE features for steering LMs. While steering can support positive use cases such as controllable text generation, personalization, and bias mitigation, it can also introduce risks that must be considered. 
In particular, steering methods can be used to manipulate model outputs in ways that circumvent safety mechanisms or amplify harmful content. 
Additionally, our methods rely on pre-trained models that may contain biases or harmful associations. 
Although our framework can help isolate and suppress such patterns, it can also be misused to reinforce them.

\section*{Acknowledgments}
This research was supported by an Azrieli Foundation Early Career Faculty Fellowship and by Open Philanthropy.
Dana Arad is supported by the Ariane de Rothschild Women Doctoral Program.
Aaron Mueller was supported by a postdoctoral fellowship under the Zuckerman STEM Leadership Program. 
This research was funded by the European Union (ERC, Control-LM, 101165402). 
Views and opinions expressed are however those of the author(s) only and do not necessarily reflect those of the European Union or the European Research Council Executive Agency.
Neither the European Union nor the granting authority can be held responsible for them.

We thank Shoval Lagziel and Yonatan Aflalo for early feedback on this work. We thank Adi Simhi for her support and feedback.

\bibliography{citations}

\appendix
\section{Additional Steering Examples}
Tables \ref{app_table:steering_examples}, \ref{app_table:pythia_steering_examples}, and \ref{app_table:llama_steering_examples} show examples of steering Gemma-2-2B, Pythia, and Llama-3.1, respectively.
Features with high output scores result in meaningful steering, while features with high input score \emph{and} low output score do not have any visible effect on the generated text.

\begin{table*}[]
% \small
\resizebox{0.95\linewidth}{!}{%
\begin{tabular}{lllllll}
\toprule
Layer & Feature & \begin{tabular}[c]{@{}l@{}}Input \\ Score\end{tabular} & \begin{tabular}[c]{@{}l@{}}Output \\ Score\end{tabular} & Top-5 Logit Lens Tokens & \begin{tabular}[c]{@{}l@{}}Optimal \\ Steering \\ Factor\end{tabular} & Generated Text \\
\midrule
11  & 10662  & \textbf{0.756} & 0   & \begin{tabular}[c]{@{}l@{}}{[}'\_engineers', '\_engineer', \\ '\_engineering', 'Engineers',\\ '\_Engineering'{]}\end{tabular} & 0.2 & \begin{tabular}[c]{@{}l@{}}Funny thing is, I bought my \\ first pair of these shoes \\ (the black leather) over 20 \\ years ago when\end{tabular}   \\ 
\\
13  & 11961  & \textbf{0.778} & 0   & \begin{tabular}[c]{@{}l@{}}{[}'\_machines', 'Machines', \\ 'machines', ' Machines', \\ '\_machine'{]}\end{tabular} & 0.2 & \begin{tabular}[c]{@{}l@{}}A friend of mine once said, "I\\ always wanted to be an architect."\end{tabular}  \\ \\
16  & 731   & \textbf{0.8}  & 0   & \begin{tabular}[c]{@{}l@{}}{[}'\_exposure', 'exposure', \\ '\_Exposure', 'Exposed', \\ '\_Exposed'{]}\end{tabular} & 0.2 & \begin{tabular}[c]{@{}l@{}}Findings show that children in the \\ United States are eating more breakfast \\ foods, but less fruit and vegetables.\end{tabular}  \\
\\
18  & 9085  & 0.023   & \textbf{0.142} & \begin{tabular}[c]{@{}l@{}}{[}' activism', ' activists', \\ ' activist', ' protest', ' \\ protesting'{]}\end{tabular}  & 6.0  & \begin{tabular}[c]{@{}l@{}}I once heard that the biggest \\
and most powerful \\ movement for human civil activism \\ and peace movement, is an \\ movement for peace movement\end{tabular} \\
\\
19  & 10015  & 0.023   & \textbf{0.734} & \begin{tabular}[c]{@{}l@{}}{[}'\_profile', '\_Profile', \\ 'Profile', 'profile', \\ '\_PROFILE'{]}\end{tabular}   & 4.0 & \begin{tabular}[c]{@{}l@{}}Findings show that a profile \\ picture in your profile is helpful \\ and makes it more likely \\ that people will add you as a\end{tabular}  \\ 
\\
19  & 10204  & 0  & \textbf{0.451} & \begin{tabular}[c]{@{}l@{}}{[}'\_crime', '\_corruption', \\ '\_violence', '\_fraud', \\ '\_crimes'{]}\end{tabular} & 1.6 & \begin{tabular}[c]{@{}l@{}}Then the man said: "If I commit murder,\\ the crime will be on my conscience."\end{tabular} 
\\
\bottomrule
\end{tabular}}
\caption{Examples of steering with features with different output and input score values in Gemma-2-2B.}
\label{app_table:steering_examples}
\end{table*}

\begin{table*}[]
% \small
\resizebox{0.95\linewidth}{!}{%
\begin{tabular}{lllllll}
\toprule
Layer & Feature & \begin{tabular}[c]{@{}l@{}}Input \\ Score\end{tabular} & \begin{tabular}[c]{@{}l@{}}Output \\ Score\end{tabular} & Top-5 Logit Lens Tokens & \begin{tabular}[c]{@{}l@{}}Optimal \\ Steering \\ Factor\end{tabular} & Generated Text \\
\midrule
4  & 7772  & 0 & \textbf{0.794}   & \begin{tabular}[c]{@{}l@{}}{[}'\_Firefox', '\_Chrome', \\ '\_browser', '\_Mozilla', \\ '\_browsers'{]}\end{tabular} & 1.2 & \begin{tabular}[c]{@{}l@{}}Findings show that a \\ site may play a key role \\ in the development of a \\ web browser.\end{tabular} \\ 
\\
5  & 21568  & 0 & \textbf{0.474}   & \begin{tabular}[c]{@{}l@{}}{[}'\_Barack', '\_Obama', \\ '\_Donald', '\_Trump', \\ '\_Bush'{]}\end{tabular} & 0.8 & \begin{tabular}[c]{@{}l@{}}The legend goes that the guy \\ was a great- Barack Obama, \\ he'd probably be the biggest \\ supporter of Barack Obama ever\end{tabular} \\ 
\bottomrule
\end{tabular}}
\caption{Examples of steering with features with different output and input score values in Pythia-70m.}
\label{app_table:pythia_steering_examples}
\end{table*}

\begin{table*}[]
% \small
\resizebox{0.95\linewidth}{!}{%
\begin{tabular}{lllllll}
\toprule
Layer & Feature & \begin{tabular}[c]{@{}l@{}}Input \\ Score\end{tabular} & \begin{tabular}[c]{@{}l@{}}Output \\ Score\end{tabular} & Top-5 Logit Lens Tokens & \begin{tabular}[c]{@{}l@{}}Optimal \\ Steering \\ Factor\end{tabular} & Generated Text \\
\midrule
2  & 25580  & \textbf{0.436} & 0  & \begin{tabular}[c]{@{}l@{}}{[}'\_sources', '\_SOUR', \\ 'sources', 'iped', \\ '/source'{]}\end{tabular} & 0.8 & \begin{tabular}[c]{@{}l@{}}That reminds me of the time \\ when I was in a public of knowledge \\ for the answer of information that said \\ the information was said that said\end{tabular}   \\ \\
20  & 17816  & 0 & \textbf{0.669}  & \begin{tabular}[c]{@{}l@{}}{[}'\_visa', '\_immigration', \\ '\_Immigration', '\_visas', \\ '\_Visa'{]}\end{tabular} & 1.6 & \begin{tabular}[c]{@{}l@{}}I believe that the H-1B visa program \\ is an important tool for employers \\ to access the best talent to fill\end{tabular}   \\ \\
26  & 21627  & 0 & \textbf{0.665}  & \begin{tabular}[c]{@{}l@{}}{[}'\_disability', '\_disabled', \\ '\_Disability', '\_disable', \\ '\_Disabled'{]}\end{tabular} & 1.2 & \begin{tabular}[c]{@{}l@{}}Findings show that the mainstream \\ media has a strong impact on public \\ opinion on disability.\end{tabular}   \\ \\
\bottomrule
\end{tabular}}
\caption{Examples of steering with features with different output and input score values in Llama-3.1-8B.}
\label{app_table:llama_steering_examples}
\end{table*}

\section{Results on Llama-3.1-8B and Pythia-70m}
\label{app:llama_pythia}
\begin{figure*}
    \centering
    \begin{subfigure}{0.4\linewidth}
        \centering
        \includegraphics[width=0.9\linewidth]{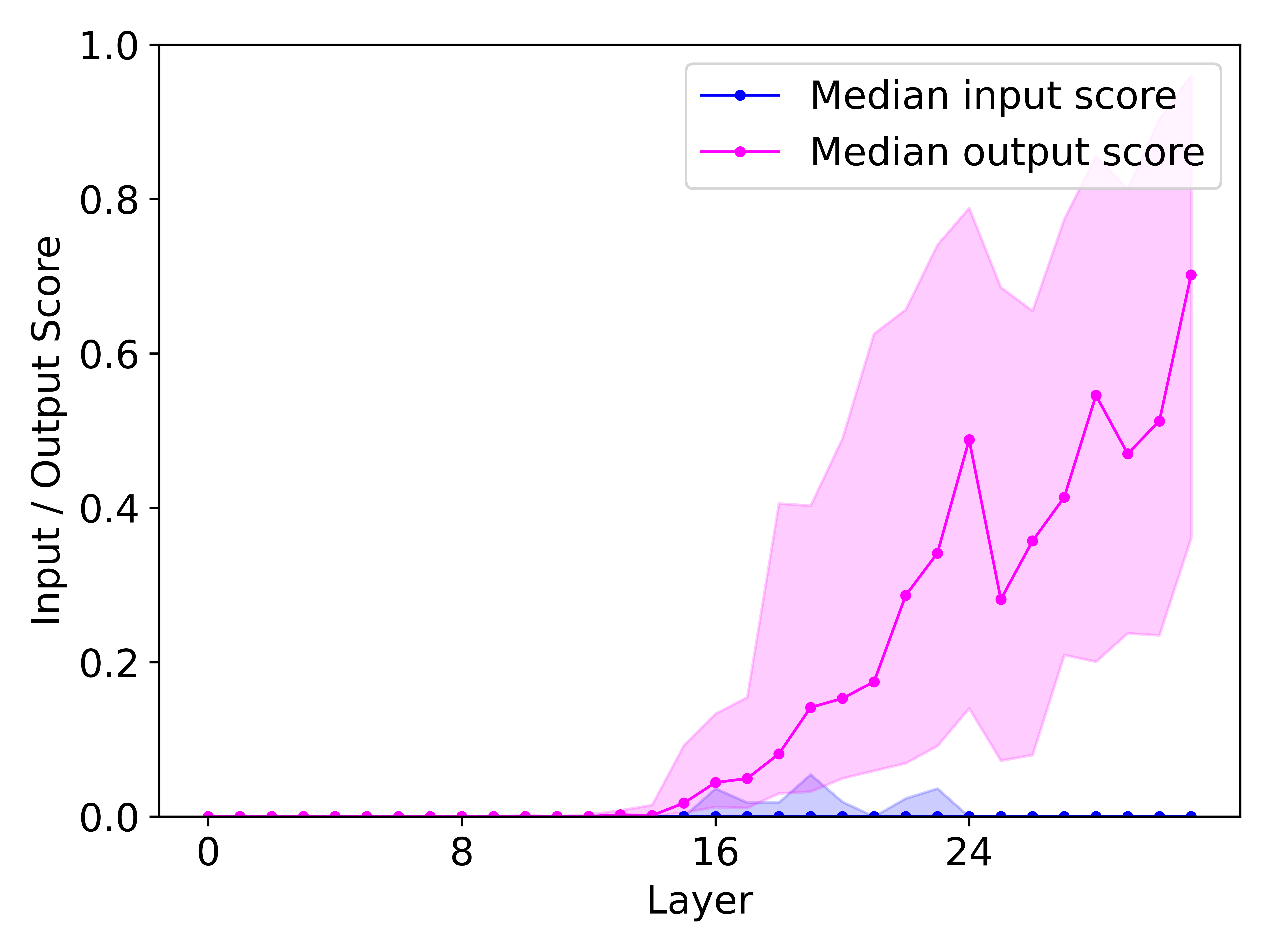}
        \caption{Llama-3.1-8B.} 
    \end{subfigure}
    \begin{subfigure}{0.4\linewidth}
        \centering
        \includegraphics[width=0.9\linewidth]{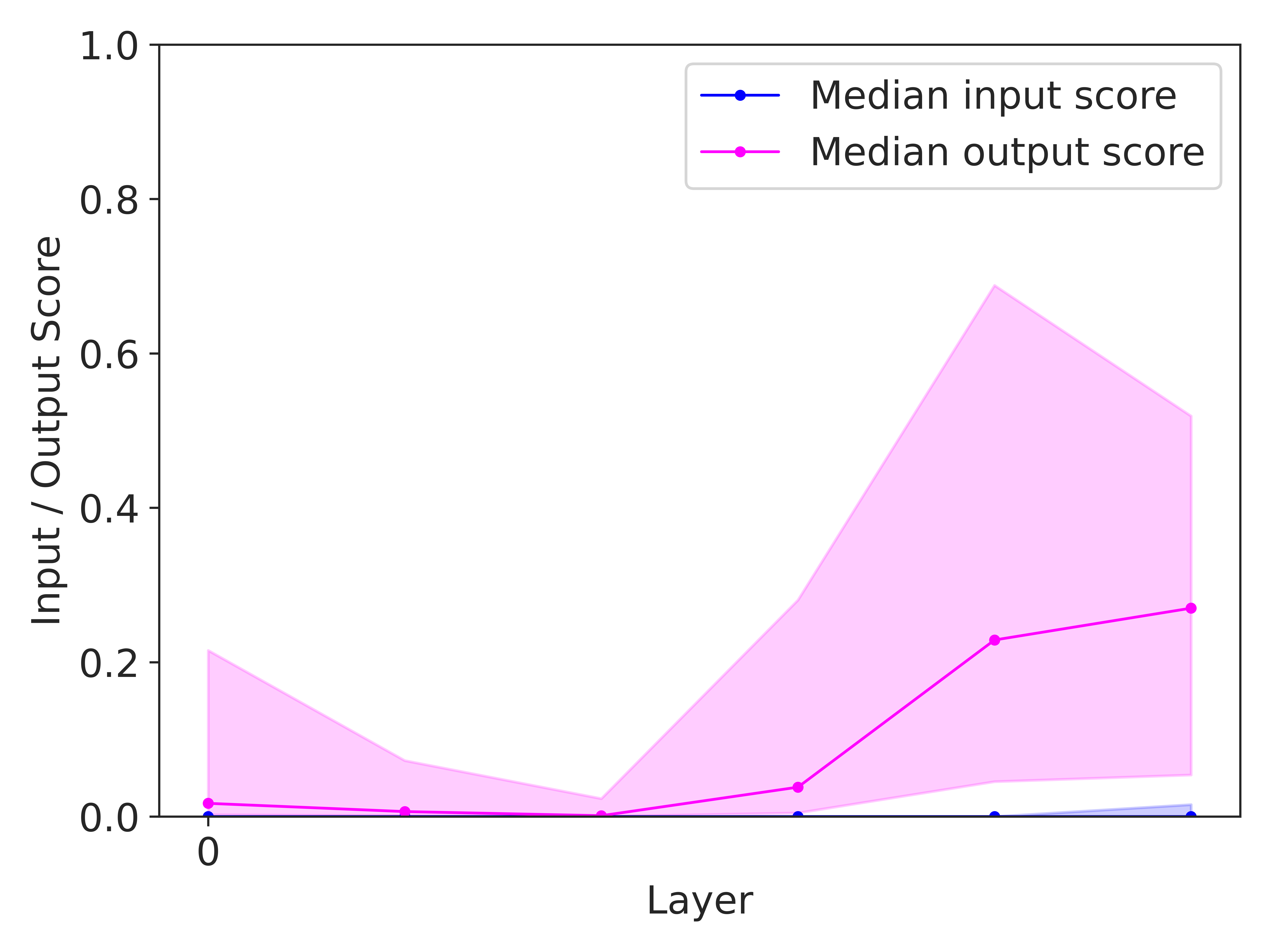}
        \caption{Pythia-70m.} 
    \end{subfigure}
    
    \caption{
        \textbf{Input and output scores across layers in Llama-3.1-8B and Pythia-70m.}
        The solid lines represent the median \textcolor{blue}{input} score (blue) and \textcolor{magenta}{output} score (magenta), while the shaded regions denote the interquartile range (25th to 75th percentile), capturing the variability across features within each layer.
        In these models we observe that high output scores emerge in later layers, while input score is mostly zero across all layers.
    }
    \label{fig:app_roles}
\end{figure*}

\Cref{fig:app_roles} demonstrates the distributions of scores across layers for Llama-3.1 and Pythia. 

\begin{figure*}
 \centering
    \begin{subfigure}{0.4\linewidth}
        \centering
        \includegraphics[width=\linewidth]{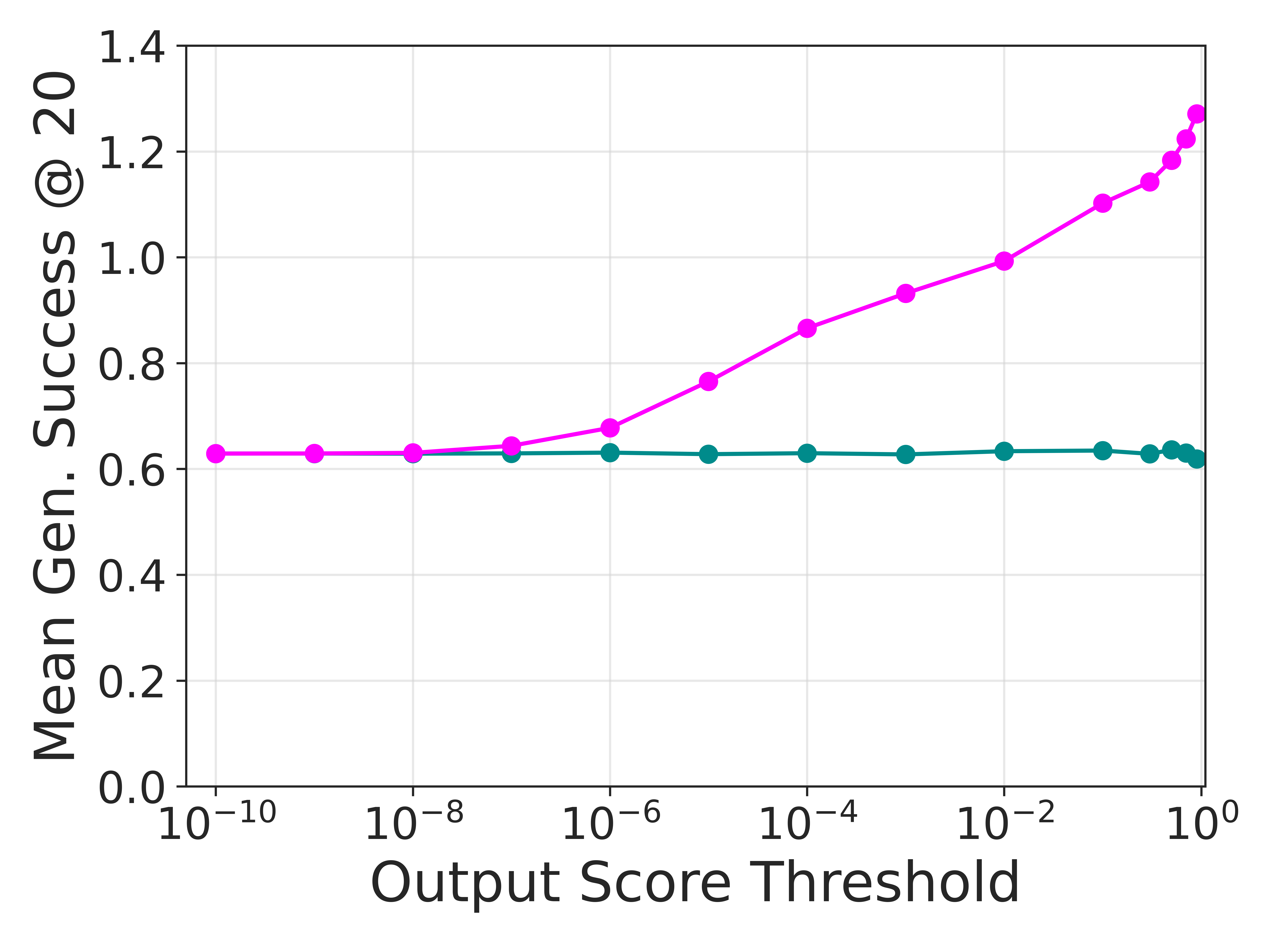}
        \caption{Llama-3.1-8B.} 
    \end{subfigure}
    \begin{subfigure}{0.4\linewidth}
        \centering
        \includegraphics[width=\linewidth]{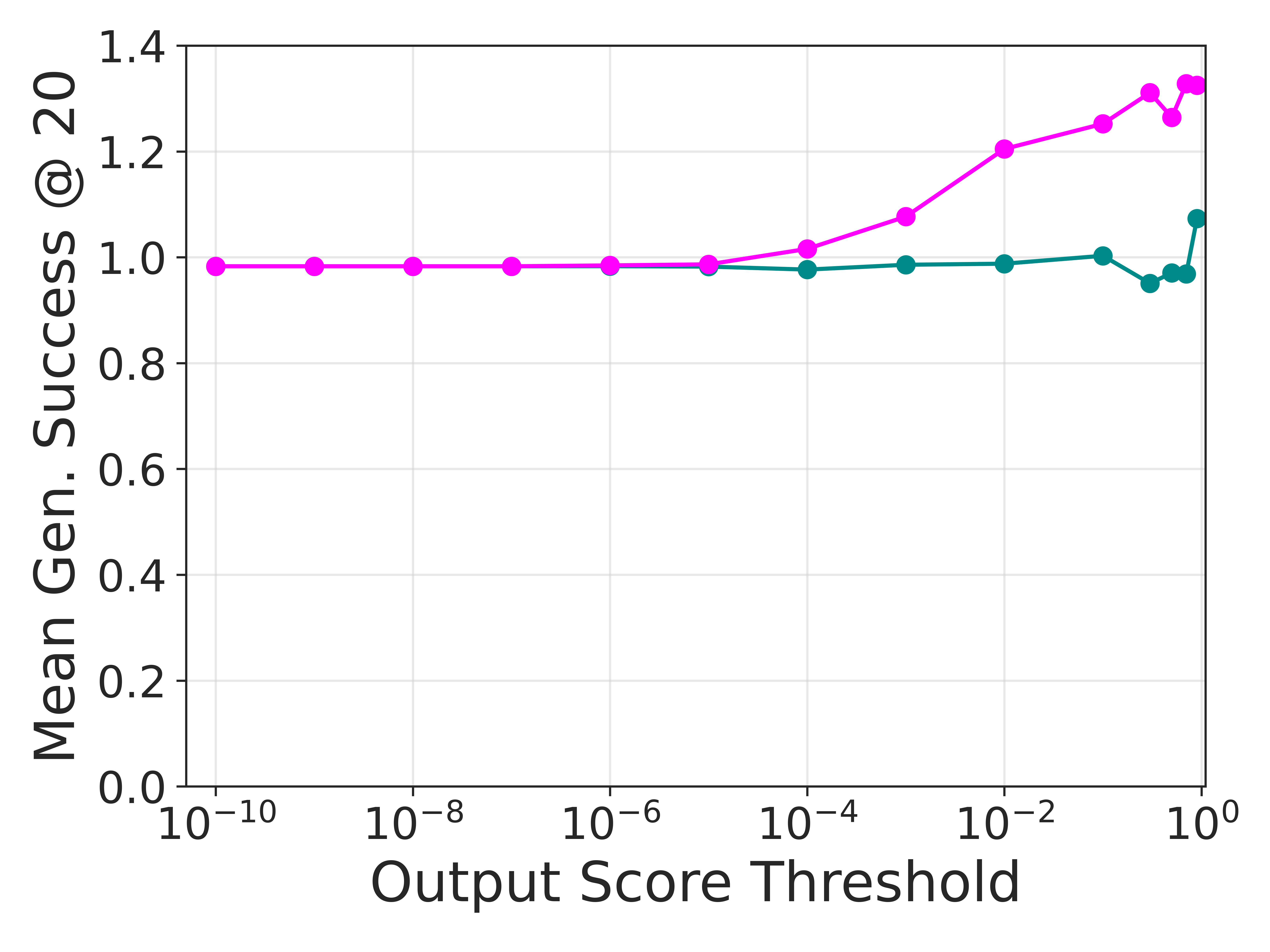}
        \caption{Pythia-70m.} 
    \end{subfigure}

 \caption{\textcolor{magenta}{Magenta} indicates the mean generation success@20 when filtering out features with output scores below different thresholds.
 \textcolor[HTML]{08A290}{Green} indicates the mean generation success@20 after filtering randomly sampled sets of features of the same size.
 Filtering results in significant increase in generation success.}
 \label{fig:app_os_th}
\end{figure*}

\Cref{fig:app_os_th} shows the mean generation success @ 20 of steered generations when filtering out features with output scores below varying thresholds (\textcolor{magenta}{magenta}) on Llama-3.1 and Pythia. 
Similarly to the Gemma models, as the threshold increases, performance improves steadily, indicating that features with higher output scores consistently lead to more successful steering. 

\section{Identifying Input Features}
\label{app:input_features}
\Cref{app_table:input_features} shows examples of features from layers 0 and 1 having high input scores. For each feature, the table includes it's top-5 logit lens tokens as well as examples for input texts that activated this feature. The tokens where the feature activated most strongly are marked using an \underline{underline}.

 \begin{table*}[]
 \small
 \label{app_table:input_features}
\begin{tabular}{lllll}
\toprule
Layer & Feature & Input Score & Top-5 Logit Lens Tokens & Activated Text \\
\midrule
0 & 725 & 0.822 & 
\begin{tabular}[c]{@{}l@{}}
{[}'\_she', '\_It', \\
'\_we', 'she', '\_OHA'{]}
\end{tabular} & 
\begin{tabular}[c]{@{}l@{}}
... euphoria and uncertainty.\underline{ He} asked himself, “ ...\\
...\underline{ She} also recalls the expeditions ...\\
... with a win.\underline{ He} said he's ...
\end{tabular} \\ \\
0 & 14772 & 0.889 & 
\begin{tabular}[c]{@{}l@{}}
{[}'inning', 'inb', 'inl',\\
'IN', 'inare'{]} 
\end{tabular} & 
\begin{tabular}[c]{@{}l@{}}
... deposition of both elast\underline{in} and fib rill\underline{in} ...\\
... I. Man\underline{in}, Three-dimensional ...\\
... $, $k\textbackslash{}\underline{in} \{1,\textbackslash{}dots ...
\end{tabular} \\ \\
1 & 4413 & 0.844 & 
\begin{tabular}[c]{@{}l@{}}
{[}'cleared', 'cle', ' clearing', \\
' clearance', ' cleaned'{]}
\end{tabular} & 
\begin{tabular}[c]{@{}l@{}}
... set. Once we\underline{ cleared} all the debris, ...\\
...\underline{ clearing} room for a future ...\\
... If this were only\underline{ cleared} away," They said ...
\end{tabular}\\ \\
1 & 3110 & 0.867 & 
\begin{tabular}[c]{@{}l@{}}
{[}'\_deltas', 'Delta', \\
'DeltaTime', \textbackslash{}Delta, 'eltas'{]}
\end{tabular} & 
\begin{tabular}[c]{@{}l@{}}
... large potential difference (\$\underline{\textbackslash{}Delta}\$ V = ...\\
... atmospheric thickness, \$\underline{\textbackslash{}Delta} z\_{eff}\$, ...\\
... finite temperature bias \$\underline{\textbackslash{}Delta} T\$ generates a ...
\end{tabular} \\
\bottomrule
\end{tabular}
 \caption{Examples of features with high input scores in early layers. Activated token are marked with an \underline{underline}.}
\end{table*}

\section{Neutral Prompt Selection}
\label{app:neutral_prompt}
Our output score relies on the use of a single neutral prompt as a prefix to the model's generation. 
To evaluate the robustness of our score to the specific choice or neutral prompt we randomly selected 10 features from each layer of Gemma-2-2B (240 features overall) and computed their output scores using all $50$ of the neutral prompts in \Cref{app_table:prefixes}.
 We find that the correlation between the output score computed with the original prompt and the mean score is $0.9557$, which is extremely high and indicates that the exact choice of the prompt has almost no impact on the results.

\section{Steering Details}
\label{app:steering}
For our main experiments, we test steering factor values $s \in \{ 0.2, 0.4, 0.8, 1.2, 1.6, 2.0, 3.0, 4.0, \\ 6.0, 8.0, 10.0, 20.0\} $.
We generate 20 tokens using a temperature of $0.7$ after each of the prefixes listed in \Cref{app_table:prefixes}.

\begin{table*}[]
\centering
\resizebox{0.95\linewidth}{!}{%
\begin{tabular}{l|l|l}
\toprule
''Findings show that'' & ''It’s no surprise that'' & ''It's been a long time since'' \\
''I once heard that'' & ''Have you ever noticed that'' & ''In my experience,'' \\
''Then the man said:'' & ''I couldn't believe when'' & ''The craziest part was when'' \\
''I believe that'' & ''The first thing I heard was'' & ''If you think about it,'' \\
''The news mentioned'' & ''Let me tell you a story about'' & ''I was shocked to learn that'' \\
''She saw a'' & ''Someone once told me that'' & ''For some reason,'' \\
''It is observed that'' & ''It might sound strange, but'' & ''I can’t help but wonder if'' \\
''Studies indicate that'' & ''They always warned me that'' & ''It makes sense that'' \\
''According to reports,'' & ''Nobody expected that'' & ''At first, I didn't believe that'' \\
''Research suggests that'' & ''Funny thing is,'' & ''That reminds me of the time when'' \\
''It has been noted that'' & ''I never thought I'd say this, but'' & ''It all comes down to'' \\
''I remember when'' & ''What surprised me most was'' & ''One time, I saw that'' \\
''It all started when'' & ''The other day, I overheard that'' & ''I was just thinking about how'' \\
''The legend goes that'' & ''Back in the day,'' & ''Imagine a world where'' \\
''If I recall correctly,'' & ''You won’t believe what happened when'' & ''They never expected that'' \\
''People often say that'' & ''A friend of mine once said,'' & ''I always knew that'' \\
''Once upon a time,'' & ''I just found out that'' & \\
\bottomrule
\end{tabular}
}
\caption{Neutral prefixes used for generation for main steering experiments.}
\label{app_table:prefixes}
\end{table*}

\section{AxBench Details}
\label{app:axbench}
\paragraph{Model Instructions.} Each concept in the Concept500 dataset is annotated as either ''text'', ''math'', or ''code''.
Following their setup we randomly sample 10 instructions per concept from instruction datasets that match the concept genre: Free Dolly dataset for text instructions \cite{DatabricksBlog2023DollyV2}, GSM8K for math \cite{cobbe2021gsm8k}, and Alpaca-Eval for code \cite{alpaca_eval}. 

\paragraph{Steering Details.} We generate up to 128 tokens per instruction, with a temperature of $0.7$, using steering factor values of $\{0.4, 0.8, 1.2, 1.6, 2.0, 3.0, 4.0, 6.0, 8.0, 10.0, 20.0, \\40.0, 60.0, 100.0\}$. 
For each concept, five instructions are used to choose the optimal steering factor (as described in \ref{subsec:steering_setup}), and the rest are used for evaluation.

\paragraph{Metrics.} We evaluate the steered texts using the metrics defined by \citeauthor{wu2025axbench}: (1) the concept score ($cs$) measures if the concept was incorporated in the generated text, (2) the fluency score ($fs$) measures the coherency of the text, and (3) the instruct score ($is$) measures the alignment of the generated text with the given instructions. 
For each metric ($m \in \{\text{cs}, \text{fs}, \text{is}\}$) and steered text $s$, an external rater returns a discrete score of either 0, 1, or 2: $m(s) \in \{0, 1, 2\}$. 
As the external LLM rater, we use Claude 3.7 sonnet (2025-02-19) \cite{claude2024}.
The prompts for all metrics are given in Tables~\ref{app_table:concept_score}, \ref{app_table:instruct_score}, and\ref{app_table:fluency_score}.

For each concept $c$, we compute each metric over the five test instructions and take the mean: $m(c) = \frac{\sum_{s \in S}{m(s)}}{|S|}$. The overall score of a concept is the harmonic mean of the three scores: $(\text{cs}(c), \text{fs}(c), \text{is}(c))$.

The cost of obtaining this score for the tested features was 65 USD.

\paragraph{Baseline Methods.} 
\Cref{table:axbench} shows the reported results of steering using various methods, as achieved by \citet{wu2025axbench}. We provide additional details on each of these methods:
\begin{itemize}
    \item \textbf{Prompt.} Given a concept, an external LLM generates a steering prompt. This prompt is then pre-pended to the actual instructions.
    \item \textbf{LoRA.}
    A parameter-efficient finetuning method \cite{xu2024lora}, trained to minimize the language modeling loss on a set of positive examples that contain the concept.
    \item \textbf{LoReFT.} Low-rank representation finetuning \cite{wu2024reft}, with the same objective as LoRA.
    \item \textbf{ReFT-r1.} A weakly supervised approach based on \citet{wu2024reft} \cite{wu2025axbench}. ReFT-r1 jointly learns concept detection and steering on supervised data by combining the training objectives of linear probing and supervised steering.
    \item \textbf{Difference-in-means.} Difference-in-means (DiffMean) \cite{pmlr-v48-larsen16} uses the difference between averaged representations from two classes of inputs as a steering vector.
    \item \textbf{SAE-A.} SAEs with AUROC selection \cite{wu2025axbench}.  For each feature, its max-pooled activations are computed over a set of positive training examples. Top-scoring features, based on the AUROC over the dataset given true labels, are used for steering.

\end{itemize}

\paragraph{License.} AxBench and Alpaca-Eval are published under an Apache license. GSM8K is published under MIT license. Dolly is published under Creative Commons Attribution Share Alike 3.0 license. All artifacts are used in compliance with their license.

\begin{table*}[]
\begin{tabular}{l}
\toprule
\textbf{Concept Score} \\
\midrule
\begin{tabular}[c]{@{}l@{}}{[}System{]}\\
Please act as an impartial judge and evaluate whether the specified concept is presented in the \\
sentence fragment provided below. Focus solely on whether the concept is clearly incorporated, \\
without regard for grammar or logical coherence. Begin your evaluation by providing a short \\
explanation of whether the concept is clearly incorporated. Point out references to the concept \\
in the fragment, noting any minimal or awkward incorporations. \\
After providing your explanation, rate the concept’s relevance on a scale \\
from 0 to 2, where 0 indicates the concept is not present at all,
1 indicates the concept is somewhat \\
present but minimally or awkwardly incorporated, and 2 indicates the concept is more fully \\ 
and effectively incorporated, with stronger and more natural integration. \\
Provide your rating using this exact format: “Rating: {[}{[}score{]}{]}”.\\ \\
{[}Concept Start{]}\\
{[}Concept goes here{]}\\
{[}Concept End{]}\\
{[}Sentence Fragment Start{]}\\
{[}Sentence goes here{]}\\
{[}Sentence Fragment End{]}\\
\bottomrule
\end{tabular}
\end{tabular}
\caption{Prompt used to compute the concept score.}
\label{app_table:concept_score}
\end{table*}

\begin{table*}[]
\centering
\begin{tabular}{l}
\toprule
\textbf{Instruct Score} \\
\midrule
\begin{tabular}[c]{@{}l@{}}{[}System{]}\\
Please act as an impartial judge and evaluate whether the sentence fragment provided below \\
is related to the instruction. Focus solely on the degree of relatedness in terms of \\
topic, regardless of grammar, coherence, or informativeness.
Begin your evaluation by \\
providing a brief explanation of whether the sentence is related to the instruction, and \\
point out references related to the instruction. After providing your explanation, rate \\
the instruction relevance on a scale from 0 to 2, where 0 indicates the sentence is unrelated \\
to the instruction, 1 indicates it is somewhat related but only minimally or indirectly \\
relevant in terms of topic, and 2 indicates it is more clearly and directly \\
related to the instruction. \\
Provide your rating using this exact format: “Rating: {[}{[}score{]}{]}”.
\\ \\
{[}Instruction Start{]}\\
{[}Instruction goes here{]}\\
{[}Instruction End{]}\\
{[}Sentence Fragment Start{]}\\
{[}Sentence goes here{]}\\
{[}Sentence Fragment End{]} \\
\end{tabular}
\\
\bottomrule
\end{tabular}
\caption{Prompt used to compute the instruct score.}
\label{app_table:instruct_score}
\end{table*}

\begin{table*}[]
\begin{tabular}{l}
\toprule
\textbf{Fluency Score} \\
\midrule
\begin{tabular}[c]{@{}l@{}}{[}System{]}\\
Please act as an impartial judge and evaluate the fluency of the sentence fragment provided \\
below. Focus solely on fluency, disregarding its completeness, relevance, coherence with any \\
broader context, or informativeness. Begin your evaluation by briefly describing the fluency \\
of the sentence, noting any unnatural phrasing, awkward transitions, grammatical errors, or \\
repetitive structures that may hinder readability. After providing your explanation, rate the \\
sentence’s fluency on a scale from 0 to 2, where 0 indicates the sentence is not fluent and \\
highly unnatural (e.g., incomprehensible or repetitive), 1 indicates it is somewhat fluent but \\
contains noticeable errors or awkward phrasing, \\
and 2 indicates the sentence is fluent and almost perfect. \\
Provide your rating using this exact format: “Rating: {[}{[}score{]}{]}”. \\ \\

{[}Sentence Fragment Start{]}\\
{[}Sentence goes here{]}\\
{[}Sentence Fragment End{]}
\end{tabular} \\
\bottomrule
\end{tabular}
\caption{Prompt used to compute the fluency score.}
\label{app_table:fluency_score} 
\end{table*}

\section{Computational Requirements}
The experiments in this work were conducted using an NVIDIA A$40$ node with $8$ $48$GB GPUs.
The complete set of experiments, including initial exploration, consumed 300-500 GPU hours.

Computing the output score for a single feature takes 6 seconds on average on a single A$40$ $48$GB GPU. Computing the input scores given pre-computed activations takes negligible time. 

\end{document}

%% file: math_commands.tex
\usepackage{amsmath,amsfonts,bm}

\def\eqref#1{equation~\ref{#1}}
\def\1{\bm{1}}

\DeclareMathAlphabet{\mathsfit}{\encodingdefault}{\sfdefault}{m}{sl}
\SetMathAlphabet{\mathsfit}{bold}{\encodingdefault}{\sfdefault}{bx}{n}

%% file: citations.bib
@article{paulo2024automatically,
  title={Automatically interpreting millions of features in large language models},
  author={Paulo, Gon{\c{c}}alo and Mallen, Alex and Juang, Caden and Belrose, Nora},
  journal={arXiv preprint arXiv:2410.13928},
  year={2024}
}

@inproceedings{makeloveevaluating,
title={Evaluating Sparse Autoencoders for Controlling Open-Ended Text Generation},
author={Aleksandar Makelov, Nathaniel Monson},
booktitle={Second NeurIPS Workshop on Attributing Model Behavior at Scale},
year={2024},
url={https://openreview.net/forum?id=8b3GqZLPoj}
}

@inproceedings{
wattenberg2024relational,
title={Relational Composition in Neural Networks: A Survey and Call to Action},
author={Martin Wattenberg and Fernanda Vi{\'e}gas},
booktitle={ICML 2024 Workshop on Mechanistic Interpretability},
year={2024},
url={https://openreview.net/forum?id=zzCEiUIPk9}
}

@inproceedings{singhvi-etal-2025-using,
    title = "Using Shapley interactions to understand how models use structure",
    author = "Singhvi, Divyansh  and
      Misra, Diganta  and
      Erkelens, Andrej  and
      Jain, Raghav  and
      Papadimitriou, Isabel  and
      Saphra, Naomi",
    editor = "Che, Wanxiang  and
      Nabende, Joyce  and
      Shutova, Ekaterina  and
      Pilehvar, Mohammad Taher",
    booktitle = "Proceedings of the 63rd Annual Meeting of the Association for Computational Linguistics (Volume 1: Long Papers)",
    month = jul,
    year = "2025",
    address = "Vienna, Austria",
    publisher = "Association for Computational Linguistics",
    url = "https://aclanthology.org/2025.acl-long.1011/",
    doi = "10.18653/v1/2025.acl-long.1011",
    pages = "20727--20737",
    ISBN = "979-8-89176-251-0"
}

@misc{nostalgebraist,
    title = {Interpreting {GPT}: The logit lens. LessWrong, 2020.},
    year = 2020,
    author = {nostalgebraist},
    url = {https://www.lesswrong.com/posts/AcKRB8wDpdaN6v6ru/interpreting-gpt-the-logit-lens}
}

@misc{saelogitlens,
    title = {Understanding SAE Features with the Logit Lens.},
    year = 2024,
    author = {Joseph Bloom and Johnny Lin},
    url = {https://www.lesswrong.com/posts/qykrYY6rXXM7EEs8Q/understanding-sae-features-with-the-logit-lens}
}

@misc{scaling2024templeton,
    title = {Scaling Monosemanticity: Extracting Interpretable Features from Claude 3 Sonnet},
    year = 2024,
    author = {Adly Templeton and Tom Conerly and Jonathan Marcus and Jack Lindsey and Trenton Bricken and Brian Chen and Adam Pearce and Craig Citro and Emmanuel Ameisen and Andy Jones and Hoagy Cunningham and Nicholas L Turner and Callum McDougall and Monte MacDiarmid and Alex Tamkin and Esin Durmus and Tristan Hume and Francesco Mosconi and C. Daniel Freeman and Theodore R. Sumers and Edward Rees and Joshua Batson and Adam Jermyn and Shan Carter and Chris Olah and Tom Henighan},
    url = {https://transformer-circuits.pub/2024/scaling-monosemanticity/}
}

@inproceedings{huang-etal-2023-rigorously,
    title = "Rigorously Assessing Natural Language Explanations of Neurons",
    author = "Huang, Jing  and
      Geiger, Atticus  and
      D{'}Oosterlinck, Karel  and
      Wu, Zhengxuan  and
      Potts, Christopher",
    editor = "Belinkov, Yonatan  and
      Hao, Sophie  and
      Jumelet, Jaap  and
      Kim, Najoung  and
      McCarthy, Arya  and
      Mohebbi, Hosein",
    booktitle = "Proceedings of the 6th BlackboxNLP Workshop: Analyzing and Interpreting Neural Networks for NLP",
    month = dec,
    year = "2023",
    address = "Singapore",
    publisher = "Association for Computational Linguistics",
    url = "https://aclanthology.org/2023.blackboxnlp-1.24/",
    doi = "10.18653/v1/2023.blackboxnlp-1.24",
    pages = "317--331"
}

@online{durmus2024steering,
author = {Esin Durmus and Alex Tamkin and Jack Clark and Jerry Wei and Jonathan Marcus and Joshua Batson and Kunal Handa and Liane Lovitt and Meg Tong and Miles McCain and Oliver Rausch and Saffron Huang and Sam Bowman and Stuart Ritchie and Tom Henighan and Deep Ganguli},
title = {Evaluating Feature Steering: A Case Study in Mitigating Social Biases},
date = {2024-10-25},
year = {2024},
url = {https://anthropic.com/research/evaluating-feature-steering},
}

@inproceedings{interpretable2024huben,
  author       = {Robert Huben and
                  Hoagy Cunningham and
                  Logan Riggs and
                  Aidan Ewart and
                  Lee Sharkey},
  title        = {Sparse Autoencoders Find Highly Interpretable Features in Language
                  Models},
  booktitle    = {The Twelfth International Conference on Learning Representations,
                  {ICLR} 2024, Vienna, Austria, May 7-11, 2024},
  publisher    = {OpenReview.net},
  year         = {2024},
  url          = {https://openreview.net/forum?id=F76bwRSLeK},
  bibsource    = {dblp computer science bibliography, https://dblp.org}
}

@article{gurarieh2025enhancing,
  author       = {Yoav Gur{-}Arieh and
                  Roy Mayan and
                  Chen Agassy and
                  Atticus Geiger and
                  Mor Geva},
  title        = {Enhancing Automated Interpretability with Output-Centric Feature Descriptions},
  journal      = {CoRR},
  volume       = {abs/2501.08319},
  year         = {2025},
  url          = {https://doi.org/10.48550/arXiv.2501.08319},
  doi          = {10.48550/ARXIV.2501.08319},
  eprinttype    = {arXiv},
  eprint       = {2501.08319},
  bibsource    = {dblp computer science bibliography, https://dblp.org}
}

@article{wu2025axbench,
  author       = {Zhengxuan Wu and
                  Aryaman Arora and
                  Atticus Geiger and
                  Zheng Wang and
                  Jing Huang and
                  Dan Jurafsky and
                  Christopher D. Manning and
                  Christopher Potts},
  title        = {AxBench: Steering LLMs? Even Simple Baselines Outperform Sparse Autoencoders},
  journal      = {CoRR},
  volume       = {abs/2501.17148},
  year         = {2025},
  url          = {https://doi.org/10.48550/arXiv.2501.17148},
  doi          = {10.48550/ARXIV.2501.17148},
  eprinttype    = {arXiv},
  eprint       = {2501.17148},
  bibsource    = {dblp computer science bibliography, https://dblp.org}
}

@article{he2024llama,
  title={Llama scope: Extracting millions of features from llama-3.1-8b with sparse autoencoders},
  author={He, Zhengfu and Shu, Wentao and Ge, Xuyang and Chen, Lingjie and Wang, Junxuan and Zhou, Yunhua and Liu, Frances and Guo, Qipeng and Huang, Xuanjing and Wu, Zuxuan and others},
  journal={arXiv preprint arXiv:2410.20526},
  year={2024}
}

@article{grattafiori2024llama,
  title={The llama 3 herd of models},
  author={Grattafiori, Aaron and Dubey, Abhimanyu and Jauhri, Abhinav and Pandey, Abhinav and Kadian, Abhishek and Al-Dahle, Ahmad and Letman, Aiesha and Mathur, Akhil and Schelten, Alan and Vaughan, Alex and others},
  journal={arXiv preprint arXiv:2407.21783},
  year={2024}
}

@inproceedings{biderman2023pythia,
  title={Pythia: A suite for analyzing large language models across training and scaling},
  author={Biderman, Stella and Schoelkopf, Hailey and Anthony, Quentin Gregory and Bradley, Herbie and O’Brien, Kyle and Hallahan, Eric and Khan, Mohammad Aflah and Purohit, Shivanshu and Prashanth, USVSN Sai and Raff, Edward and others},
  booktitle={International Conference on Machine Learning},
  pages={2397--2430},
  year={2023},
  organization={PMLR}
}

@misc{neuronpedia,
    title = {Neuronpedia: Interactive Reference and Tooling for Analyzing Neural Networks},
    year = {2023},
    note = {Software available from neuronpedia.org},
    url = {https://www.neuronpedia.org},
    author = {Lin, Johnny}
}

@article{team2024gemma,
  title={Gemma 2: Improving open language models at a practical size},
  author={Team, Gemma and Riviere, Morgane and Pathak, Shreya and Sessa, Pier Giuseppe and Hardin, Cassidy and Bhupatiraju, Surya and Hussenot, L{\'e}onard and Mesnard, Thomas and Shahriari, Bobak and Ram{\'e}, Alexandre and others},
  journal={arXiv preprint arXiv:2408.00118},
  year={2024}
}

@article{lieberum2024gemma,
  title={Gemma scope: Open sparse autoencoders everywhere all at once on gemma 2},
  author={Lieberum, Tom and Rajamanoharan, Senthooran and Conmy, Arthur and Smith, Lewis and Sonnerat, Nicolas and Varma, Vikrant and Kram{\'a}r, J{\'a}nos and Dragan, Anca and Shah, Rohin and Nanda, Neel},
  journal={arXiv preprint arXiv:2408.05147},
  year={2024}
}

@article{bricken2023monosemanticity,
   title={Towards Monosemanticity: Decomposing Language Models With Dictionary Learning},
   author={Bricken, Trenton and Templeton, Adly and Batson, Joshua and Chen, Brian and Jermyn, Adam and Conerly, Tom and Turner, Nick and Anil, Cem and Denison, Carson and Askell, Amanda and Lasenby, Robert and Wu, Yifan and Kravec, Shauna and Schiefer, Nicholas and Maxwell, Tim and Joseph, Nicholas and Hatfield-Dodds, Zac and Tamkin, Alex and Nguyen, Karina and McLean, Brayden and Burke, Josiah E and Hume, Tristan and Carter, Shan and Henighan, Tom and Olah, Christopher},
   year={2023},
   journal={Transformer Circuits Thread},
   note={https://transformer-circuits.pub/2023/monosemantic-features/index.html}
}

@inproceedings{katz2024backward,
  author       = {Shahar Katz and
                  Yonatan Belinkov and
                  Mor Geva and
                  Lior Wolf},
  editor       = {Yaser Al{-}Onaizan and
                  Mohit Bansal and
                  Yun{-}Nung Chen},
  title        = {Backward Lens: Projecting Language Model Gradients into the Vocabulary
                  Space},
  booktitle    = {Proceedings of the 2024 Conference on Empirical Methods in Natural
                  Language Processing, {EMNLP} 2024, Miami, FL, USA, November 12-16,
                  2024},
  pages        = {2390--2422},
  publisher    = {Association for Computational Linguistics},
  year         = {2024},
  url          = {https://aclanthology.org/2024.emnlp-main.142},
  bibsource    = {dblp computer science bibliography, https://dblp.org}
}

@inproceedings{dar2023analyzing,
  author       = {Guy Dar and
                  Mor Geva and
                  Ankit Gupta and
                  Jonathan Berant},
  editor       = {Anna Rogers and
                  Jordan L. Boyd{-}Graber and
                  Naoaki Okazaki},
  title        = {Analyzing Transformers in Embedding Space},
  booktitle    = {Proceedings of the 61st Annual Meeting of the Association for Computational
                  Linguistics (Volume 1: Long Papers), {ACL} 2023, Toronto, Canada,
                  July 9-14, 2023},
  pages        = {16124--16170},
  publisher    = {Association for Computational Linguistics},
  year         = {2023},
  url          = {https://doi.org/10.18653/v1/2023.acl-long.893},
  doi          = {10.18653/V1/2023.ACL-LONG.893},
  bibsource    = {dblp computer science bibliography, https://dblp.org}
}

@inproceedings{toker2024lens,
  author       = {Michael Toker and
                  Hadas Orgad and
                  Mor Ventura and
                  Dana Arad and
                  Yonatan Belinkov},
  editor       = {Lun{-}Wei Ku and
                  Andre Martins and
                  Vivek Srikumar},
  title        = {Diffusion Lens: Interpreting Text Encoders in Text-to-Image Pipelines},
  booktitle    = {Proceedings of the 62nd Annual Meeting of the Association for Computational
                  Linguistics (Volume 1: Long Papers), {ACL} 2024, Bangkok, Thailand,
                  August 11-16, 2024},
  pages        = {9713--9728},
  publisher    = {Association for Computational Linguistics},
  year         = {2024},
  url          = {https://doi.org/10.18653/v1/2024.acl-long.524},
  doi          = {10.18653/V1/2024.ACL-LONG.524},
  bibsource    = {dblp computer science bibliography, https://dblp.org}
}

@online{DatabricksBlog2023DollyV2,
    author    = {Mike Conover and Matt Hayes and Ankit Mathur and Jianwei Xie and Jun Wan and Sam Shah and Ali Ghodsi and Patrick Wendell and Matei Zaharia and Reynold Xin},
    title     = {Free Dolly: Introducing the World's First Truly Open Instruction-Tuned LLM},
    year      = {2023},
    url       = {https://www.databricks.com/blog/2023/04/12/dolly-first-open-commercially-viable-instruction-tuned-llm},
    urldate   = {2023-06-30}
}

@article{cobbe2021gsm8k,
  title={Training Verifiers to Solve Math Word Problems},
  author={Cobbe, Karl and Kosaraju, Vineet and Bavarian, Mohammad and Chen, Mark and Jun, Heewoo and Kaiser, Lukasz and Plappert, Matthias and Tworek, Jerry and Hilton, Jacob and Nakano, Reiichiro and Hesse, Christopher and Schulman, John},
  journal={arXiv preprint arXiv:2110.14168},
  year={2021}
}

@misc{alpaca_eval,
  author = {Xuechen Li and Tianyi Zhang and Yann Dubois and Rohan Taori and Ishaan Gulrajani and Carlos Guestrin and Percy Liang and Tatsunori B. Hashimoto },
  title = {AlpacaEval: An Automatic Evaluator of Instruction-following Models},
  year = {2023},
  month = {5},
  publisher = {GitHub},
  journal = {GitHub repository},
  howpublished = {\url{https://github.com/tatsu-lab/alpaca_eval}}
}

@article{elazar2021amnesic,
  title={Amnesic probing: Behavioral explanation with amnesic counterfactuals},
  author={Elazar, Yanai and Ravfogel, Shauli and Jacovi, Alon and Goldberg, Yoav},
  journal={Transactions of the Association for Computational Linguistics},
  volume={9},
  pages={160--175},
  year={2021},
  publisher={MIT Press One Rogers Street, Cambridge, MA 02142-1209, USA journals-info~…}
}

@inproceedings{bert2019tenney,
  author       = {Ian Tenney and
                  Dipanjan Das and
                  Ellie Pavlick},
  editor       = {Anna Korhonen and
                  David R. Traum and
                  Llu{\'{\i}}s M{\`{a}}rquez},
  title        = {{BERT} Rediscovers the Classical {NLP} Pipeline},
  booktitle    = {Proceedings of the 57th Conference of the Association for Computational
                  Linguistics, {ACL} 2019, Florence, Italy, July 28- August 2, 2019,
                  Volume 1: Long Papers},
  pages        = {4593--4601},
  publisher    = {Association for Computational Linguistics},
  year         = {2019},
  url          = {https://doi.org/10.18653/v1/p19-1452},
  doi          = {10.18653/V1/P19-1452},
  bibsource    = {dblp computer science bibliography, https://dblp.org}
}

@inproceedings{meng2022locating,
  author       = {Kevin Meng and
                  David Bau and
                  Alex Andonian and
                  Yonatan Belinkov},
  editor       = {Sanmi Koyejo and
                  S. Mohamed and
                  A. Agarwal and
                  Danielle Belgrave and
                  K. Cho and
                  A. Oh},
  title        = {Locating and Editing Factual Associations in {GPT}},
  booktitle    = {Advances in Neural Information Processing Systems 35: Annual Conference
                  on Neural Information Processing Systems 2022, NeurIPS 2022, New Orleans,
                  LA, USA, November 28 - December 9, 2022},
  year         = {2022},
  url          = {http://papers.nips.cc/paper\_files/paper/2022/hash/6f1d43d5a82a37e89b0665b33bf3a182-Abstract-Conference.html},
  bibsource    = {dblp computer science bibliography, https://dblp.org}
}

@inproceedings{brunner2020identifiability,
  author       = {Gino Brunner and
                  Yang Liu and
                  Damian Pascual and
                  Oliver Richter and
                  Massimiliano Ciaramita and
                  Roger Wattenhofer},
  title        = {On Identifiability in Transformers},
  booktitle    = {8th International Conference on Learning Representations, {ICLR} 2020,
                  Addis Ababa, Ethiopia, April 26-30, 2020},
  publisher    = {OpenReview.net},
  year         = {2020},
  url          = {https://openreview.net/forum?id=BJg1f6EFDB},
  bibsource    = {dblp computer science bibliography, https://dblp.org}
}

@inproceedings{geva2021keyvalue,
  author       = {Mor Geva and
                  Roei Schuster and
                  Jonathan Berant and
                  Omer Levy},
  editor       = {Marie{-}Francine Moens and
                  Xuanjing Huang and
                  Lucia Specia and
                  Scott Wen{-}tau Yih},
  title        = {Transformer Feed-Forward Layers Are Key-Value Memories},
  booktitle    = {Proceedings of the 2021 Conference on Empirical Methods in Natural
                  Language Processing, {EMNLP} 2021, Virtual Event / Punta Cana, Dominican
                  Republic, 7-11 November, 2021},
  pages        = {5484--5495},
  publisher    = {Association for Computational Linguistics},
  year         = {2021},
  url          = {https://doi.org/10.18653/v1/2021.emnlp-main.446},
  doi          = {10.18653/V1/2021.EMNLP-MAIN.446},
  bibsource    = {dblp computer science bibliography, https://dblp.org}
}

@inproceedings{hernandez2024linearity,
  author       = {Evan Hernandez and
                  Arnab Sen Sharma and
                  Tal Haklay and
                  Kevin Meng and
                  Martin Wattenberg and
                  Jacob Andreas and
                  Yonatan Belinkov and
                  David Bau},
  title        = {Linearity of Relation Decoding in Transformer Language Models},
  booktitle    = {The Twelfth International Conference on Learning Representations,
                  {ICLR} 2024, Vienna, Austria, May 7-11, 2024},
  publisher    = {OpenReview.net},
  year         = {2024},
  url          = {https://openreview.net/forum?id=w7LU2s14kE},
  bibsource    = {dblp computer science bibliography, https://dblp.org}
}

@inproceedings{geva2023recall,
  author       = {Mor Geva and
                  Jasmijn Bastings and
                  Katja Filippova and
                  Amir Globerson},
  editor       = {Houda Bouamor and
                  Juan Pino and
                  Kalika Bali},
  title        = {Dissecting Recall of Factual Associations in Auto-Regressive Language
                  Models},
  booktitle    = {Proceedings of the 2023 Conference on Empirical Methods in Natural
                  Language Processing, {EMNLP} 2023, Singapore, December 6-10, 2023},
  pages        = {12216--12235},
  publisher    = {Association for Computational Linguistics},
  year         = {2023},
  url          = {https://doi.org/10.18653/v1/2023.emnlp-main.751},
  doi          = {10.18653/V1/2023.EMNLP-MAIN.751},
  bibsource    = {dblp computer science bibliography, https://dblp.org}
}

@inproceedings{liu2019linguistic,
  author       = {Nelson F. Liu and
                  Matt Gardner and
                  Yonatan Belinkov and
                  Matthew E. Peters and
                  Noah A. Smith},
  editor       = {Jill Burstein and
                  Christy Doran and
                  Thamar Solorio},
  title        = {Linguistic Knowledge and Transferability of Contextual Representations},
  booktitle    = {Proceedings of the 2019 Conference of the North American Chapter of
                  the Association for Computational Linguistics: Human Language Technologies,
                  {NAACL-HLT} 2019, Minneapolis, MN, USA, June 2-7, 2019, Volume 1 (Long
                  and Short Papers)},
  pages        = {1073--1094},
  publisher    = {Association for Computational Linguistics},
  year         = {2019},
  url          = {https://doi.org/10.18653/v1/n19-1112},
  doi          = {10.18653/V1/N19-1112},
  bibsource    = {dblp computer science bibliography, https://dblp.org}
}

@inproceedings{belinkov2017translation,
  author       = {Yonatan Belinkov and
                  Nadir Durrani and
                  Fahim Dalvi and
                  Hassan Sajjad and
                  James R. Glass},
  editor       = {Regina Barzilay and
                  Min{-}Yen Kan},
  title        = {What do Neural Machine Translation Models Learn about Morphology?},
  booktitle    = {Proceedings of the 55th Annual Meeting of the Association for Computational
                  Linguistics, {ACL} 2017, Vancouver, Canada, July 30 - August 4, Volume
                  1: Long Papers},
  pages        = {861--872},
  publisher    = {Association for Computational Linguistics},
  year         = {2017},
  url          = {https://doi.org/10.18653/v1/P17-1080},
  doi          = {10.18653/V1/P17-1080},
  bibsource    = {dblp computer science bibliography, https://dblp.org}
}

@article{zhang2018language,
  title={Language modeling teaches you more syntax than translation does: Lessons learned through auxiliary task analysis},
  author={Zhang, Kelly W and Bowman, Samuel R},
  journal={arXiv preprint arXiv:1809.10040},
  year={2018}
}

@article{lad2024mstages,
  author       = {Vedang Lad and
                  Wes Gurnee and
                  Max Tegmark},
  title        = {The Remarkable Robustness of LLMs: Stages of Inference?},
  journal      = {CoRR},
  volume       = {abs/2406.19384},
  year         = {2024},
  url          = {https://doi.org/10.48550/arXiv.2406.19384},
  doi          = {10.48550/ARXIV.2406.19384},
  eprinttype    = {arXiv},
  eprint       = {2406.19384},
  bibsource    = {dblp computer science bibliography, https://dblp.org}
}

@inproceedings{
marks2025sparse,
title={Sparse Feature Circuits: Discovering and Editing Interpretable Causal Graphs in Language Models},
author={Samuel Marks and Can Rager and Eric J Michaud and Yonatan Belinkov and David Bau and Aaron Mueller},
booktitle={The Thirteenth International Conference on Learning Representations},
year={2025},
url={https://openreview.net/forum?id=I4e82CIDxv}
}

@article{gurnee2024universal,
  author       = {Wes Gurnee and
                  Theo Horsley and
                  Zifan Carl Guo and
                  Tara Rezaei Kheirkhah and
                  Qinyi Sun and
                  Will Hathaway and
                  Neel Nanda and
                  Dimitris Bertsimas},
  title        = {Universal Neurons in {GPT2} Language Models},
  journal      = {Trans. Mach. Learn. Res.},
  volume       = {2024},
  year         = {2024},
  url          = {https://openreview.net/forum?id=ZeI104QZ8I},
  bibsource    = {dblp computer science bibliography, https://dblp.org}
}

@inproceedings{bird-loper-2004-nltk,
    title = "{NLTK}: The Natural Language Toolkit",
    author = "Bird, Steven  and
      Loper, Edward",
    booktitle = "Proceedings of the {ACL} Interactive Poster and Demonstration Sessions",
    month = jul,
    year = "2004",
    address = "Barcelona, Spain",
    publisher = "Association for Computational Linguistics",
    url = "https://aclanthology.org/P04-3031/",
    pages = "214--217"
}

@misc{claude2024,
  author       = {Anthropic},
  title        = {The Claude 3 Model Family: Opus, Sonnet, Haiku},
  year         = {2024},
  howpublished = {\url{https://assets.anthropic.com/m/61e7d27f8c8f5919/original/Claude-3-Model-Card.pdf}},
  note         = {Accessed: 2025-04}
}

@misc{bills2024explaining,
  author       = {Steven Bills and Nick Cammarata and Dan Mossing and Henk Tillman and Leo Gao and Gabriel Goh and Ilya Sutskever and Jan Leike and Jeff Wu and William Saunders},
  title        = {Language models can explain neurons in language models},
  year         = {2023},
  howpublished = {\url{https://openaipublic.blob.core.windows.net/neuron-explainer/paper/index.html}}
}

@article{zou2023representation,
  title={Representation engineering: A top-down approach to ai transparency},
  author={Zou, Andy and Phan, Long and Chen, Sarah and Campbell, James and Guo, Phillip and Ren, Richard and Pan, Alexander and Yin, Xuwang and Mazeika, Mantas and Dombrowski, Ann-Kathrin and others},
  journal={arXiv preprint arXiv:2310.01405},
  year={2023}
}

@article{li2023inference,
  title={Inference-time intervention: Eliciting truthful answers from a language model},
  author={Li, Kenneth and Patel, Oam and Vi{\'e}gas, Fernanda and Pfister, Hanspeter and Wattenberg, Martin},
  journal={Advances in Neural Information Processing Systems},
  volume={36},
  pages={41451--41530},
  year={2023}
}

@article{simhi2024constructing,
  title={Constructing benchmarks and interventions for combating hallucinations in llms},
  author={Simhi, Adi and Herzig, Jonathan and Szpektor, Idan and Belinkov, Yonatan},
  journal={arXiv preprint arXiv:2404.09971},
  year={2024}
}

@inproceedings{teehan2022emergent,
  title={Emergent structures and training dynamics in large language models},
  author={Teehan, Ryan and Clinciu, Miruna and Serikov, Oleg and Szczechla, Eliza and Seelam, Natasha and Mirkin, Shachar and Gokaslan, Aaron},
  booktitle={Proceedings of BigScience Episode\# 5--Workshop on Challenges \& Perspectives in Creating Large Language Models},
  pages={146--159},
  year={2022}
}

@article{turner2023steering,
  title={Steering language models with activation engineering},
  author={Turner, Alexander Matt and Thiergart, Lisa and Leech, Gavin and Udell, David and Vazquez, Juan J and Mini, Ulisse and MacDiarmid, Monte},
  journal={arXiv preprint arXiv:2308.10248},
  year={2023}
}

@article{liu2023context,
  title={In-context vectors: Making in context learning more effective and controllable through latent space steering},
  author={Liu, Sheng and Ye, Haotian and Xing, Lei and Zou, James},
  journal={arXiv preprint arXiv:2311.06668},
  year={2023}
}

@article{van2024extending,
  title={Extending activation steering to broad skills and multiple behaviours},
  author={van der Weij, Teun and Poesio, Massimo and Schoots, Nandi},
  journal={arXiv preprint arXiv:2403.05767},
  year={2024}
}

@inproceedings{rimsky2024steering,
  title={Steering Llama 2 via Contrastive Activation Addition},
  author={Rimsky, Nina and Gabrieli, Nick and Schulz, Julian and Tong, Meg and Hubinger, Evan and Turner, Alexander},
  booktitle={Proceedings of the 62nd Annual Meeting of the Association for Computational Linguistics (Volume 1: Long Papers)},
  pages={15504--15522},
  year={2024}
}

@inproceedings{lai2024style,
  title={Style-Specific Neurons for Steering LLMs in Text Style Transfer},
  author={Lai, Wen and Hangya, Viktor and Fraser, Alexander},
  booktitle={Proceedings of the 2024 Conference on Empirical Methods in Natural Language Processing},
  pages={13427--13443},
  year={2024}
}

@InProceedings{pmlr-v48-larsen16,
  title = 	 {Autoencoding beyond pixels using a learned similarity metric},
  author = 	 {Larsen, Anders Boesen Lindbo and Sønderby, Søren Kaae and Larochelle, Hugo and Winther, Ole},
  booktitle = 	 {Proceedings of The 33rd International Conference on Machine Learning},
  pages = 	 {1558--1566},
  year = 	 {2016},
  editor = 	 {Balcan, Maria Florina and Weinberger, Kilian Q.},
  volume = 	 {48},
  series = 	 {Proceedings of Machine Learning Research},
  address = 	 {New York, New York, USA},
  month = 	 {20--22 Jun},
  publisher =    {PMLR},
  url = 	 {https://proceedings.mlr.press/v48/larsen16.html}
}

@inproceedings{subramani2022extracting,
  title={Extracting Latent Steering Vectors from Pretrained Language Models},
  author={Subramani, Nishant and Suresh, Nivedita and Peters, Matthew E},
  booktitle={Findings of the Association for Computational Linguistics: ACL 2022},
  pages={566--581},
  year={2022}
}

@article{o2024steering,
  title={Steering language model refusal with sparse autoencoders},
  author={O'Brien, Kyle and Majercak, David and Fernandes, Xavier and Edgar, Richard and Chen, Jingya and Nori, Harsha and Carignan, Dean and Horvitz, Eric and Poursabzi-Sangde, Forough},
  journal={arXiv preprint arXiv:2411.11296},
  year={2024}
}

@inproceedings{taveekitworachai2024null,
  title={Null-shot prompting: rethinking prompting large language models with hallucination},
  author={Taveekitworachai, Pittawat and Abdullah, Febri and Thawonmas, Ruck},
  booktitle={Proceedings of the 2024 Conference on Empirical Methods in Natural Language Processing},
  pages={13321--13361},
  year={2024}
}

@inproceedings{bau2018identifying,
title={Identifying and Controlling Important Neurons in Neural Machine Translation},
author={Anthony Bau and Yonatan Belinkov and Hassan Sajjad and Nadir Durrani and Fahim Dalvi and James Glass},
booktitle={International Conference on Learning Representations},
year={2019},
url={https://openreview.net/forum?id=H1z-PsR5KX},
}

@inproceedings{xu2024lora,
  author       = {Yuhui Xu and
                  Lingxi Xie and
                  Xiaotao Gu and
                  Xin Chen and
                  Heng Chang and
                  Hengheng Zhang and
                  Zhengsu Chen and
                  Xiaopeng Zhang and
                  Qi Tian},
  title        = {QA-LoRA: Quantization-Aware Low-Rank Adaptation of Large Language
                  Models},
  booktitle    = {The Twelfth International Conference on Learning Representations,
                  {ICLR} 2024, Vienna, Austria, May 7-11, 2024},
  publisher    = {OpenReview.net},
  year         = {2024},
  url          = {https://openreview.net/forum?id=WvFoJccpo8},
  bibsource    = {dblp computer science bibliography, https://dblp.org}
}

@article{rajamanoharan2024jumprelu,
  author       = {Senthooran Rajamanoharan and
                  Tom Lieberum and
                  Nicolas Sonnerat and
                  Arthur Conmy and
                  Vikrant Varma and
                  J{\'{a}}nos Kram{\'{a}}r and
                  Neel Nanda},
  title        = {Jumping Ahead: Improving Reconstruction Fidelity with JumpReLU Sparse
                  Autoencoders},
  journal      = {CoRR},
  volume       = {abs/2407.14435},
  year         = {2024},
  url          = {https://doi.org/10.48550/arXiv.2407.14435},
  doi          = {10.48550/ARXIV.2407.14435},
  eprinttype    = {arXiv},
  eprint       = {2407.14435},
  bibsource    = {dblp computer science bibliography, https://dblp.org}
}

@article{wu2024reft,
  title={Reft: Representation finetuning for language models},
  author={Wu, Zhengxuan and Arora, Aryaman and Wang, Zheng and Geiger, Atticus and Jurafsky, Dan and Manning, Christopher D and Potts, Christopher},
  journal={Advances in Neural Information Processing Systems},
  volume={37},
  pages={63908--63962},
  year={2024}
}

@article{li2025fairsteer,
  title={FairSteer: Inference Time Debiasing for LLMs with Dynamic Activation Steering},
  author={Li, Yichen and Fan, Zhiting and Chen, Ruizhe and Gai, Xiaotang and Gong, Luqi and Zhang, Yan and Liu, Zuozhu},
  journal={arXiv preprint arXiv:2504.14492},
  year={2025}
}

@inproceedings{arad2024refact,
  title={ReFACT: Updating Text-to-Image Models by Editing the Text Encoder},
  author={Arad, Dana and Orgad, Hadas and Belinkov, Yonatan},
  booktitle={Proceedings of the 2024 Conference of the North American Chapter of the Association for Computational Linguistics: Human Language Technologies (Volume 1: Long Papers)},
  pages={2537--2558},
  year={2024}
}
